\documentclass[10pt]{article}
\usepackage[T1]{fontenc}
\usepackage[utf8]{inputenc}
\usepackage[a4paper,textwidth=132mm,hmarginratio=1:1,top=26mm,bottom=30mm,footskip=14mm]{geometry}

\usepackage[lining]{PlayfairDisplay}
\usepackage[semibold,lining,proportional]{sourceserifpro}
\usepackage[semibold,lining,proportional]{sourcesanspro}
\usepackage[scale=0.86]{sourcecodepro}

\makeatletter
\renewcommand{\normalsize}{\@setfontsize\normalsize{10.4}{12}\abovedisplayskip 9\p@ plus2\p@ minus4\p@\abovedisplayshortskip \z@ plus2\p@\belowdisplayshortskip 5\p@ plus2\p@ minus2\p@\belowdisplayskip \abovedisplayskip}
\makeatother
\AtBeginDocument{\normalsize}
\usepackage{amsmath,amssymb}
\usepackage{newtxmath}
\usepackage[activate={true,nocompatibility},final,tracking=false,factor=1100,stretch=12,shrink=12]{microtype}

\usepackage{xcolor}
\definecolor{ink}{HTML}{13235A}
\definecolor{cobalt}{HTML}{2548BC}
\definecolor{coral}{HTML}{D9553A}
\definecolor{muted}{HTML}{5C6585}
\definecolor{hair}{HTML}{D5D8E3}
\definecolor{paper}{HTML}{F4F1EA}
\definecolor{wash}{HTML}{EEF1FB}

\usepackage{graphicx}
\usepackage{booktabs,array,colortbl,tabularx}
\usepackage{alltt}
\usepackage{enumitem}
\usepackage{etoolbox}
\usepackage{titlesec}
\usepackage{fancyhdr}
\usepackage[hang,flushmargin,bottom]{footmisc}
\usepackage[labelfont={bf,color=cobalt},textfont={color=ink},font={footnotesize,sf},labelsep=quad,justification=raggedright,singlelinecheck=false,skip=6pt]{caption}
\usepackage[most]{tcolorbox}
\usepackage{xurl}
\usepackage[round,sort,authoryear]{natbib}
\usepackage{hyperref}
\hypersetup{colorlinks=true,linkcolor=cobalt,citecolor=cobalt,urlcolor=cobalt,pdfborder={0 0 0},
  pdfdisplaydoctitle=true,bookmarksnumbered=true}

\setlist{leftmargin=1.2em,itemsep=0.28em,topsep=0.3em,parsep=0pt,partopsep=0pt}
\setlist[itemize]{label={\color{cobalt}\textbullet}}
\titleformat{\section}{\sffamily\fontsize{15.5}{19}\selectfont\bfseries\color{ink}}{\llap{\textcolor{cobalt}{\thesection}\hspace{3.2mm}}}{0pt}{}
\titleformat{\subsection}{\sffamily\fontsize{11}{14}\selectfont\bfseries\color{ink}}{\llap{\textcolor{muted}{\mdseries\small\thesubsection}\hspace{3.2mm}}}{0pt}{}
\titleformat{\paragraph}[runin]{\sffamily\normalsize\bfseries\color{ink}}{}{0pt}{}[\enspace]
\titlespacing*{\section}{0pt}{2.1em plus 0.3em minus 0.2em}{0.55em}
\titlespacing*{\subsection}{0pt}{1.25em plus 0.2em minus 0.1em}{0.28em}
\titlespacing*{\paragraph}{0pt}{0.75em}{0pt}

\newcommand{\standfirst}[1]{{\par\nobreak\sffamily\fontsize{11}{14.6}\selectfont\interlinepenalty=10000\leavevmode\color{ink}#1\par\vspace{0.5em}}}

\fancypagestyle{plain}{\fancyhf{}\fancyfoot[R]{\sffamily\scriptsize\color{muted}\thepage}}

\newlength{\figwidth}
\newcommand{\widefig}[2][\figwidth]{\makebox[\textwidth][c]{\includegraphics[width=#1]{figures/#2}}}
\newcommand{\us}{\rowcolor{wash}}
\renewcommand{\arraystretch}{1.16}
\arrayrulecolor{hair}
\newcolumntype{R}{>{\raggedleft\arraybackslash}p}
\newcommand{\tabfont}{\sffamily\footnotesize\color{ink}}
\newcommand{\thd}[1]{{\color{muted}#1}}
\newcommand{\toprl}{\arrayrulecolor{ink}\toprule\arrayrulecolor{hair}}
\newcommand{\botrl}{\arrayrulecolor{ink}\bottomrule\arrayrulecolor{hair}}

\newtcolorbox{lesson}{enhanced,frame hidden,boxrule=0pt,sharp corners,colback=paper,left=9pt,right=9pt,top=6pt,bottom=7pt,
  borderline west={1.6pt}{0pt}{cobalt},before skip=0.9em,after skip=0.9em,fontupper=\sffamily\small\color{ink},parbox=false,before upper={\setlength{\parskip}{0.4em}}}

\newtcolorbox{terms}{enhanced,frame hidden,boxrule=0pt,sharp corners,colback=wash,left=9pt,right=9pt,top=7pt,bottom=7pt,before skip=1em,after skip=1em,fontupper=\sffamily\footnotesize\color{ink}}

\newtcolorbox{codeblock}{enhanced,frame hidden,boxrule=0pt,sharp corners,colback=paper,left=8pt,right=8pt,top=5pt,bottom=5pt,before skip=0.6em,after skip=0.8em,fontupper=\ttfamily\footnotesize\color{ink}}
\newcommand{\code}[1]{{\ttfamily\small #1}}
\newcommand{\tr}[1]{\textit{#1}}  

\newcommand{\statnum}[1]{{\sffamily\fontsize{17}{19}\selectfont\color{ink}#1}}
\newcommand{\statlab}[1]{{\sffamily\scriptsize\color{muted}#1}}

\hypersetup{pdftitle={Antalia-2 Mini: A 7.6-million-parameter Turkish text-to-speech model trained entirely on synthetic speech},
  pdfauthor={Sezgin Saygili, Emre Kaplaner, Oncel Ozgul, Fikri San Koktas},
  pdfsubject={Text-to-speech},pdfkeywords={text-to-speech, Turkish, flow matching, synthetic training data}}

\begin{document}
\thispagestyle{plain}
\begingroup
\setlength{\parskip}{0pt}
{\sffamily\scriptsize\color{muted}\textls[70]{\MakeUppercase{Technical report}}\hfill\textls[40]{November 2026}\par}
\vspace{5.5mm}
{\playfairLF\fontsize{34}{37}\selectfont\color{ink}Antalia-2 \textit{\color{cobalt}Mini}\par}
\vspace{2.4mm}
{\sffamily\fontsize{13.2}{17}\selectfont\color{ink}A 7.6-million-parameter Turkish text-to-speech model\\ trained entirely on synthetic speech\par}
\vspace{4mm}
{\sffamily\small\bfseries\color{ink}Sezgin Saygili\hspace{1.05em}Emre Kaplaner\hspace{1.05em}Oncel Ozgul\hspace{1.05em}Fikri San Koktas\par}
\vspace{0.9mm}
{\sffamily\footnotesize\color{muted}PatientDesk AI\hspace{1em}\href{mailto:sezgin@patientdesk.ai}{\color{muted}sezgin@patientdesk.ai}\par}
\vspace{3.4mm}
{\color{hair}\rule{\textwidth}{0.4pt}\par}
\vspace{2.2mm}
\noindent\begin{tabular*}{\textwidth}{@{}l@{\extracolsep{\fill}}l@{\extracolsep{\fill}}l@{\extracolsep{\fill}}l@{\extracolsep{\fill}}l@{}}
\statnum{7.62M}&\statnum{15.3\,MB}&\statnum{48\,kHz}&\statnum{0.22\%}&\statnum{41\texttimes}\\[0.6mm]
\statlab{parameters}&\statlab{optional fp16 weights}&\statlab{audio output}&\statlab{character error rate, public harness}&\statlab{real time, M5 CPU, 2 threads}
\end{tabular*}\par
\vspace{1.8mm}
{\color{hair}\rule{\textwidth}{0.4pt}\par}
\vspace{1.6mm}
{\sffamily\scriptsize\color{muted}\href{https://huggingface.co/cloud0day3/antalia-mini}{Model and weights}\hspace{1.3em}\href{https://huggingface.co/spaces/cloud0day3/antalia-mini}{Browser demo}\hspace{1.3em}\href{https://pypi.org/project/antalia-mini/}{\texttt{pip install antalia-mini}}\hspace{1.3em}\href{https://speech.patientdesk.ai/en/benchmarks}{Benchmark}\hspace{1.3em}Apache-2.0\par}
\vspace{3.6mm}
\endgroup

{\fontsize{9.9}{11.7}\selectfont{\sffamily\bfseries\small\color{cobalt}Abstract}\enspace Antalia-2 Mini is a single-voice Turkish text-to-speech model with 7,620,931 parameters. It reads characters and produces 48\,kHz audio at 41 times real time on an Apple M5 laptop CPU (2 threads). Its training audio came only from VoxCPM2 (the teacher), about 300 times larger, cloning one synthetic clip; automatic checks kept about 1,790\,h. Fine-tuning the vocoder on the acoustic model's own mel spectrograms cut the word error rate (WER) of Whisper's transcripts from a mean of 3.89\% to 1.75\% at adjacent pre-release checkpoints. The teacher corrupts the start of a clip when the first word begins with \tr{ö}, the student learned this, and a spoken prefix we cut from the output hides it. On the 495-sentence Freya-TR-Eval benchmark, where we tuned settings, version 1.0.1 has WER 1.09\% and character error rate 0.22\% on our public harness. A paired bootstrap cannot distinguish these from EMA Lightning, an 8.6M Turkish model, which scores higher on full-band UTMOS, an English-trained quality proxy. No human listening test exists. Weights, inference code and the benchmark harness are public under Apache-2.0; training code and data are not.\par}

{\setlength{\intextsep}{7pt plus 1pt}
\begin{figure}[!h]
\widefig{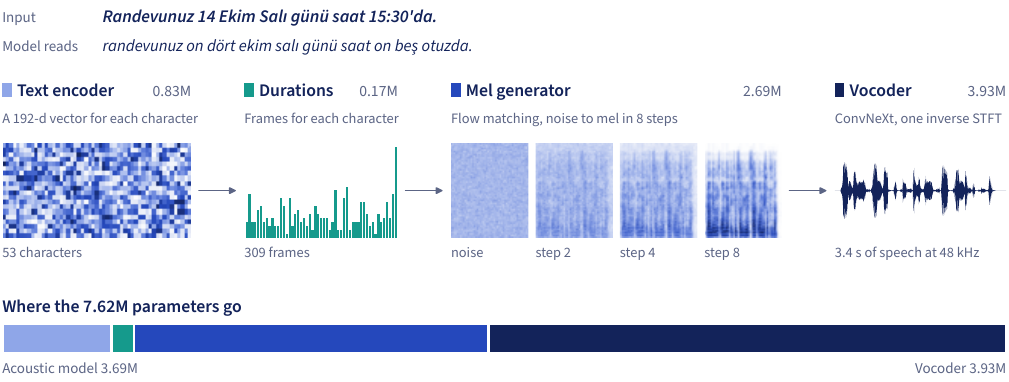}
\caption{\textbf{The whole model on one real sentence (version 1.0.1, seed 0).} The normalizer writes the numbers out, the encoder and duration predictor plan the timing, the generator draws a mel spectrogram from noise in 8 steps, and the vocoder turns it into audio.}\label{fig:hero}
\end{figure}}

\section{Introduction}\label{sec:intro}
\standfirst{Antalia-2 Mini has 7.62M parameters, reads characters and learned only from the audio of a model about 300 times larger. On one benchmark a paired bootstrap cannot distinguish its word and character error rates from that teacher's or EMA Lightning's, and no human listening test of it exists yet.}

A text-to-speech (TTS) model small enough for a CPU can run on the machine that needs the speech, with no GPU server and no text leaving the device. Among the Turkish systems we ran, two small open-weights models, Piper (voice dfki) at 16M parameters and MMS-TTS at 36M, have word error rates (WER) of 3.41\% and 6.38\%, against 1.16\% for Trendyol-TTS at about 2.38B. We get a WER by having the Whisper speech recognizer \citep{radford2022whisper} transcribe a model's audio and comparing the transcript with the input text, here on the public harness of the Freya-TR-Eval benchmark. Antalia-2 Mini (Mini below) aims for the low error rates of large systems at 7.62M, in one voice.

Its default weights are 30.5\,MB at fp32, with an optional 15.3\,MB fp16 download (Table~\ref{tab:artifacts}). A browser page and a signed Android engine also run it on the device, but no real phone has been measured.

Mini reads characters, 47 symbols in all, with no phonemizer to turn spelling into sound symbols, and learns how letters sound from its training audio. Turkish spelling suits this choice. \citet{oney1997beginning} describe it as phonologically transparent, with regular letter-sound correspondences, and \citet{babayigit2022symmetrically} names \tr{Ğ} as the exception. A text normalizer first turns numbers, dates, times, amounts, initialisms, web addresses and English or brand words into spoken form (Table~\ref{tab:normalizer}), and its known defects carry into the speech.

All of Mini's training audio came from one model, VoxCPM2 \citep{zhou2026voxcpm2}, roughly 300 to 313 times its size. We had this teacher read text of our choosing in one synthetic voice, so every clip came with the text it was asked to read. Mini, the student, trains on those clips with no access to the teacher's internal features. Section~\ref{sec:related} traces this idea back at least to TTS-by-TTS \citep{hwang2021ttsbytts}. Teacher errors become training targets, as \citet{pipatanakul2026thai} state, and Section~\ref{sec:fixes} documents one that Mini learned.

Among the other systems we ran, EMA Lightning \citep{aslan2026emalightning} is nearest ours in size and recipe. Its card says it was also trained only on synthetic speech from VoxCPM2, about 1,000\,h, so the two models share a teacher model, with different voices. Our scoring protocol follows that card. The vocoder is the network that turns a mel spectrogram into audio, and the acoustic model before it turns text into that mel. Mini has 10.93\% fewer parameters, 7,620,931 against 8,556,172, with the vocoder in both counts.

\begin{figure}[t]
\widefig{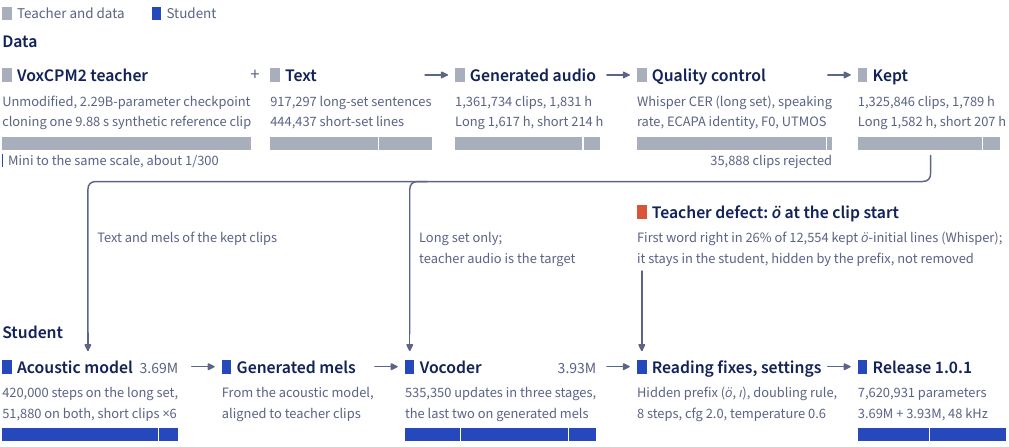}
\caption{\textbf{All training audio came from unmodified VoxCPM2 cloning one synthetic clip, and the vocoder was fine-tuned on the acoustic model's own mels.} Bars are proportional within each node; generated and kept hours share one scale. F0 is pitch, ECAPA identity is a speaker embedding's similarity to the voice, and cfg is classifier-free guidance, a sampler setting. Coral shows a Whisper rate on kept long-set lines.}\label{fig:recipe}
\end{figure}

We report six results, and Figure~\ref{fig:recipe} traces the recipe behind them. The terms box defines the words they use.

\begin{figure}[t]
\begin{terms}
\begin{tabularx}{\linewidth}{@{}>{\bfseries\raggedright\arraybackslash}p{21mm}>{\raggedright\arraybackslash}X@{}}
\multicolumn{2}{@{}l}{\bfseries\small Terms used in this report}\\[2pt]
Mel & A mel spectrogram, audio as energy per frequency band over time, with bands spaced to follow human pitch perception. Ours has 128 bands and 93.75 frames per second. Real mels come from the teacher's audio, generated mels from the acoustic model.\\
Copy synthesis & Computing a mel from existing audio, here the teacher's, and passing it through the vocoder alone. It shows what the vocoder loses by itself.\\\midrule
Teacher, student & The teacher, unmodified VoxCPM2, generated all training audio. The student, Antalia-2 Mini, trained on its clips.\\
Long set, short set & Training clips of sentences up to about 25\,s, and of short utterances, mostly fragments of 2 to 6 words.\\
Runs & Acoustic model: the main run on the long set, then the short-clip fine-tune on both sets; a sibling run on filtered clips was stopped. Vocoder: the real-mel vocoder, then the first and final vocoder fine-tunes on generated mels (Table~\ref{tab:checkpoints}).\\
Versions 1.0.0 and 1.0.1 & Two public releases with identical weights. Version 1.0.1 changes only the default output chain, the fixed signal processing after the vocoder.\\
Final settings & 8 sampling steps, guidance 2.0, temperature 0.6, speed 0.95, hidden prefix and suffix, doubling rule 1.3 (Sections~\ref{sec:model} and~\ref{sec:fixes}).\\
Alania, Antalia & Our model families: Alania models are closed and sold through an API, and Antalia models have open weights.\\\midrule
Take & One render of all 495 benchmark sentences.\\
Public harness & Our test setup for Freya-TR-Eval with public code: 3 takes, seed 100000\,\textit{s}\,+\,\textit{i} for sentence \textit{i} of take \textit{s}, one NVIDIA L4 each for Mini and EMA Lightning (different VMs).\\
Own harness & Our second setup, with the same scoring and private code: 3 takes, each with one seed for all its sentences (seeds 0, 1, 2). Mini 1.0.1 ran them on an L40S, an A10G and an A100, EMA Lightning on an H200, and the teacher 1 seed on an A100.\\
Band matching & Audio is resampled down to 8\,kHz and back up to 16\,kHz before Whisper hears it, as the benchmark's card recommends.\\
WER, CER & Word and character error rate: word or character edit distance between Whisper's transcript and the reference, over reference length, in percent. On the benchmark, Whisper (large-v3, beam 5) hears band-matched audio and each take gets one score.\\
Points & Percentage points, our unit for differences: 1.09\% against 1.04\% is 0.05 points.\\
Not distinguishable & The paired 95\% interval of a difference includes 0. Figures also say not separable or within noise. ``Level with'' is a stricter test (Section~\ref{sec:eval}).\\
One seed's span & About $\pm$0.13 WER points, the standard deviation across Mini 1.0.0's 3 own-harness seeds. The $\pm$0.27 half-width of Section~\ref{sec:eval} comes from resampling sentences instead.\\
Clean rate & The share of sentences with no word error in a take, averaged over takes.\\
UTMOS & UTMOS22-strong, an automatic estimate of the mean opinion score (MOS), the average rating listeners give, trained on English speech; higher is better. For scale, the teacher scores 3.42 on the benchmark (full band, own harness, 1 seed). It runs on 16\,kHz audio, so neither UTMOS nor WER hears anything above 8\,kHz. Full-band UTMOS skips band matching and 8\,kHz UTMOS includes it.\\\midrule
RTF, times real time & Real-time factor is wall-clock synthesis time over audio duration. Times real time is its inverse, so RTF 0.024 is 41\texttimes{}.\\
\end{tabularx}
\end{terms}
\end{figure}

\begin{itemize}
\item On the 495 sentences of Freya-TR-Eval \citep{pamuk2026freyatts}, Mini 1.0.1 has WER 1.09\% and CER 0.220\% on the public harness. A paired 95\% bootstrap distinguishes it from neither EMA Lightning 1.0.1 nor, on our own harness, the teacher (1 seed), and both score higher than Mini on full-band UTMOS, an English-trained proxy. We tuned Mini's settings on these sentences, so final-settings numbers are optimistic by selection (Section~\ref{sec:eval}).
\item Fine-tuning the vocoder on generated mels cut WER from a mean of 3.89\% to 1.75\% at adjacent pre-release checkpoints (7 pairs, same seed, our own harness), while a frozen vocoder stayed at 3.63--4.14\% (Section~\ref{sec:vocoder}).
\item By Whisper, the teacher gets first words right on 26\% of 12,554 kept \tr{ö}-initial long-set lines, against 94\% for other capital first letters. The student learned this defect, and a spoken prefix cut from the output hides it without removing it (Section~\ref{sec:fixes}).
\item A mixed-precision weight cache (bf16 autocast) silently gave the generator, the part that draws the mel, no gradients for the first 6,000 steps of the main run (Section~\ref{sec:training}).
\item One machine with 8 GPUs generated all 1,361,734 clips. Automatic checks kept about 1,790\,h, but Whisper checked only 7,500 of the 444,437 short clips (Section~\ref{sec:data}).
\item Mini 1.0.1 runs at 35\texttimes{} real time on one NVIDIA L4 with 8 vCPUs (all 495 sentences), against 58\texttimes{} for EMA Lightning 1.0.1 in plain PyTorch and 331\texttimes{} compiled, with 16 vCPUs. Mini 1.0.0 was 1.36 times faster than plain EMA Lightning 1.0.1 on one 8-vCPU AMD EPYC (first 100 sentences), and 1.0.1 is untimed there (Section~\ref{sec:speed}).
\end{itemize}

\section{The model}\label{sec:model}

\standfirst{An acoustic model reads characters, its generator draws a mel spectrogram in 8 steps, and a vocoder turns the mel into 48\,kHz audio in one pass. At the default settings the generator, trained with shortcut self-consistency, does 84.0\% of the multiply-accumulates, and the model also renders hidden text that is cut from the output.}

Table~\ref{tab:parts} lists the model's 7,620,931 parameters at inference.

\begin{table}[t]
\caption{\textbf{The vocoder holds 51.62\% of the parameters and does 14.4\% of the operations; the generator holds 35.24\% and does 84.0\%.} Shipped parts are counted from the headers of the released weight files. Operations are multiply-accumulates at the defaults on a 5.11\,s sentence, from forward hooks on linear and convolution layers and a formula for attention, without LayerNorm, activations, the inverse Fourier transform or the output chain. The Generator row's 84.0\% leaves out its condition ConvNeXt, which does a further 1.1\%. Counts of the pieces are in parentheses. No part has dropout layers. dBFS is decibels relative to digital full scale, and Q sets a filter's bandwidth.}\label{tab:parts}
\centering\tabfont
\makebox[\textwidth][c]{\begin{tabular}{@{}l>{\raggedright\arraybackslash}p{95mm}rrr@{}}\toprl
\thd{Part} & \leavevmode\thd{Built from} & \thd{Parameters} & \thd{Share} & \thd{Operations}\\\midrule
Text encoder & Embedding 47\,\texttimes\,192 (9,024); 3 ConvNeXt blocks, hidden 384, kernel 5 (449,280); attention block, 4 heads (369,408); LayerNorm (384) & 828,096 & 10.87\% & 0.5\%\\
Duration predictor & Convolution from 192 to 128; 3 convolutions of width 128, kernel 3; final convolution & 173,441 & 2.28\% & 0.1\%\\
Generator & 5 DiT blocks, width 192, 4 heads, SwiGLU hidden 512 (2,217,600); condition ConvNeXt, hidden 384, kernel 7 (150,144); adaLN-single projection (148,608); input projection (86,208); output projection (49,408); time MLP (32,896); final table (384); null condition (192) & 2,685,440 & 35.24\% & 84.0\%\\
\textbf{Acoustic model} & The three parts above & \textbf{3,686,977} & \textbf{48.38\%} & \\
\textbf{Vocoder} & Input convolution from 128 to 256, kernel 7 (229,632); 8 ConvNeXt blocks, width 256, hidden 768, kernel 7 (3,176,448); head to 2,050 values (526,850); 2~LayerNorms (512 each). Receptive field 55 frames, 587\,ms & \textbf{3,933,954} & \textbf{51.62\%} & \textbf{14.4\%}\\
\us\textbf{Total at inference} & & \textbf{7,620,931} & \textbf{100\%} & \textbf{100\%}\\\midrule
\multicolumn{5}{@{}l}{\thd{Training only, not shipped}}\\
MAS prior & Linear layer from the encoder's output to the mel bands & 24,704 & & \\
Discriminators & 5 period, at periods 2, 3, 5, 7, 11 (41,105,770); 4 spectrogram, at FFT sizes 512, 1,024, 2,048, 4,096 (203,528) & 41,309,298 & & \\\midrule
\multicolumn{5}{@{}l}{\thd{Fixed signal processing in version 1.0.1, no learned parameters}}\\
Equalizer & High-pass at 80\,Hz, Q 0.707; low shelf of $-$4\,dB at 200\,Hz; peaks of $-$2.5\,dB at 350\,Hz, Q 1.2, and +3\,dB at 3.2\,kHz, Q 0.9; high shelf of +2\,dB at 8\,kHz. Runs as one 1,744-tap convolution at 48\,kHz & & & \\
De-esser & Acts on the band above 5.5\,kHz; threshold $-$34\,dBFS, ratio 3, attack 1.5\,ms, release 60\,ms; off at 8\,kHz output & & & \\
Gain, limiter & Fixed +1.6\,dB; tanh knee above $-$1\,dBFS, ceiling $-$0.1\,dBFS. No per-utterance loudness normalization and, by default, no silence trimming & & & \\
Output rates & 48\,kHz native; 24, 16 and 8\,kHz by Kaiser-windowed sinc resampling & & & \\\botrl
\end{tabular}}
\end{table}

\paragraph{Text input.} The acoustic model reads 47 symbols: padding, the BOS and EOS markers for the start and end of a sequence, the space, 8 punctuation marks, 32 letters (29 Turkish plus q, w and x) and 3 circumflex vowels. It has no phonemes and no digits, so a normalizer inside the package first writes everything else in these symbols (Table~\ref{tab:normalizer}). Our checklist records that no native speaker has reviewed address readings such as \tr{dabılyu} and \tr{bölü}.

\begin{table}[t]
\caption{\textbf{The normalizer writes digits, amounts, initialisms and addresses as words, and it has known defects.} Real calls of \code{Antalia.normalize}, package 1.0.1. The right column is the exact string returned, before the hidden prefix and suffix added at synthesis. A word counts as English when it is in the 126,052-entry CMU pronouncing dictionary \protect\citep{cmudict}, a runtime dependency, and not in a list of 225,764 words seen in Turkish text.}\label{tab:normalizer}
\centering\tabfont
\makebox[\textwidth][c]{\begin{tabular}{@{}>{\raggedright\arraybackslash}p{32mm}>{\raggedright\arraybackslash}p{57mm}>{\raggedright\arraybackslash}p{72mm}@{}}\toprl
\leavevmode\thd{Case} & \leavevmode\thd{Input} & \leavevmode\thd{Model text}\\\midrule
Date, time, title & Randevunuz 14 Ekim Salı günü saat 15:30'da, Dr. Ayşe Yılmaz ile. & randevunuz on dört ekim salı günü saat on beş otuzda, doktor ayşe yılmaz ile.\\
Area, plus sign, amount & Oda 45 m² ve 3+1, fiyatı 2,5 milyon TL. & oda kırk beş metrekare ve üç artı bir, fiyatı iki buçuk milyon lira.\\
Dose, blood pressure & Günde 2x1 tablet, 500 mg; tansiyonunuz 120/80 mmHg. & günde iki defa bir tablet, beş yüz miligram; tansiyonunuz yüz yirmiye seksen milimetre cıva.\\
Initialisms with suffixes & SGK ve KDV kesintileri için THY'nin CEO'su TBMM'de konuştu. & segeka ve kadeve kesintileri için teheye'nin siio'su tebememe'de konuştu.\\
Brand names & PatientDesk'e WhatsApp ya da ChatGPT üzerinden yazabilirsiniz. & peyşınt desk'e whatsapp ya da çet ci pi ti üzerinden yazabilirsiniz.\\
Web address & Ayrıntılar için www.ornek.com.tr/kampanya adresini ziyaret edin. & ayrıntılar için dabılyu dabılyu dabılyu nokta ornek nokta kom nokta te re bölü kampanya adresini ziyaret edin.\\\midrule
\multicolumn{3}{@{}l}{\thd{Known defects seen in real calls}}\\
Suffix \tr{-ki} takes the wrong vowel (should be \tr{otuzdaki}) & Saat 10:30'daki toplantı iptal. & saat on otuzdakı toplantı iptal.\\
Range with currency keeps the hyphen & Fiyat 500-750 TL arası. & fiyat beş yüz-yedi yüz elli lira arası.\\
Unit mg/dL not handled & Kan şekeri 110 mg/dL çıktı. & kan şekeri yüz on miligram dl çıktı.\\
Words seen in Turkish text are never respelled & Outlook, Notion ve podcast uygulamaları. & outlook, notion ve podcast uygulamaları.\\\botrl
\end{tabular}}
\end{table}

\paragraph{Audio features.} A mel frame covers 2,048 samples of 48\,kHz audio. Frames start 512 samples apart (10.667\,ms), and the bands follow the Slaney mel scale from 0 to 24\,kHz. We normalize as $y = (\ln\max(\mathrm{mel},10^{-5}) + 5.9)/2.6$. The generator stacks frames in pairs, so a token lasts 21.333\,ms and holds 256 values.

\paragraph{Encoder, alignment, durations.} The text encoder is an embedding followed by 3 ConvNeXt blocks \citep{liu2022convnext}, a kind of convolutional residual block, and one self-attention block with rotary position embeddings, or RoPE \citep{su2024roformer}.

Nothing in a training clip says which frames belong to which character. Monotonic alignment search, MAS \citep{kim2020glowtts}, finds this. A training-only layer, the prior, maps each encoder output to an expected mel frame, and a dynamic program picks the most likely left-to-right assignment, with at least one frame per symbol. A small convolutional duration predictor learns the resulting lengths, with gradients stopped at the encoder output, and replaces MAS at synthesis.

\paragraph{Flow matching.} The generator follows a learned velocity from noise at $t = 0$ to a mel at $t = 1$. Flow matching \citep{lipman2023flow} trains that velocity by regression. We mix a target mel $x_1$ with Gaussian noise $x_0$, and the network, given the text condition $c$ and a step size of 0 (an input the next paragraph explains), predicts the velocity of the straight line between them:
\begin{equation}
x_t = (1-t)\,x_0 + t\,x_1, \qquad
\mathcal{L}_\text{FM} = \bigl\| v_\theta(x_t, t, 0 \mid c) - (x_1 - x_0) \bigr\|^2 .
\end{equation}

\begin{figure}[t]
\widefig[132mm]{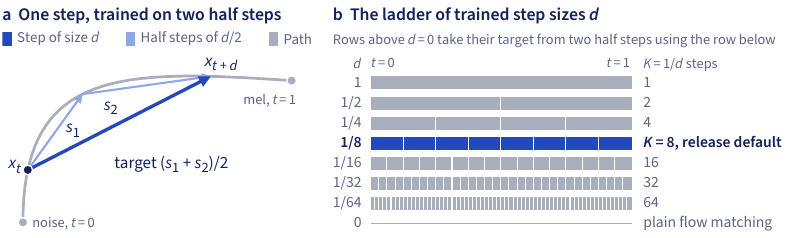}
\caption{\textbf{Shortcut self-consistency training lets one model sample at 1, 2, 4 or 8 steps; the release uses 8, and the benchmark scored 4, 8 and 16 only.} (a) Schematic, after \protect\citet{frans2025shortcut}. (b) Rows are to scale on \textit{t}. At the final settings on our own harness, 4 steps gave WER 1.17 against 0.97 at 8 (3 seeds; Section~\ref{sec:eval}).}\label{fig:shortcut}
\end{figure}

\paragraph{Shortcut self-consistency.} The learned velocity averages over many such lines, so its paths curve. Small steps follow them but cost many network passes, and a large step leaves them. Shortcut self-consistency \citep{frans2025shortcut} gives the network the step size $d$ and trains one step of size $d$ to land where two steps of size $d/2$ land (Figure~\ref{fig:shortcut}a). The half steps run on the current weights with no gradients, and one call at the full size regresses onto the mean of their velocities:
\begin{align}
s_1 &= v_\theta\bigl(x_t,\, t,\, \tfrac{d}{2} \mid c\bigr), \qquad
s_2 = v_\theta\bigl(x_t + \tfrac{d}{2}\,s_1,\; t + \tfrac{d}{2},\; \tfrac{d}{2} \mid c\bigr), \nonumber\\
\mathcal{L}_\text{SC} &= \bigl\| v_\theta(x_t, t, d \mid c) - \tfrac{1}{2}(s_1 + s_2) \bigr\|^2 .
\end{align}

Each utterance trained this way draws one of 7 levels uniformly, from $d = 1$ down to $d = 1/64$ (Figure~\ref{fig:shortcut}b). A level takes its half steps from the level below, and the last from plain flow matching at $d = 0$. One model can then sample at 1, 2, 4 or 8 steps, and the other powers of 2 up to 64 also lie on the trained grid. On the benchmark we scored 4, 8 and 16 steps only.

Training sums these two losses and one term each for the prior and the duration predictor, with unit weights. A quarter of each batch trains $\mathcal{L}_\text{SC}$ and the rest $\mathcal{L}_\text{FM}$. Each is divided by its own token count, so both carry equal total weight.

\paragraph{Generator network.} The generator is a diffusion transformer, DiT \citep{peebles2023dit}, of 5 blocks with RoPE and gated SwiGLU feed-forward layers \citep{shazeer2020glu}. The encoder output, repeated per frame, passed through one ConvNeXt block and averaged per token, joins $x_t$ at the input. Time and step size enter through adaptive layer normalization (adaLN), where a small network sets each normalization layer's scale and shift. In adaLN-single \citep{chen2024pixart} all blocks share that network, and each block adds a small learned table.

\paragraph{Sampling.} Classifier-free guidance \citep{ho2022cfg} runs the generator twice per step, with the text condition and with a learned null vector that replaces the condition on 10\% of training utterances. The sampler extrapolates past the conditional velocity $v_c$, away from the null velocity $v_\varnothing$, with weight $w$, and takes $K$ Euler steps from noise scaled by a temperature of 0.6:
\begin{equation}
\hat v = v_c + (w-1)(v_c - v_\varnothing), \qquad
x^{i+1} = x^{i} + \tfrac{1}{K}\,\hat v\bigl(x^{i},\, \tfrac{i}{K},\, \tfrac{1}{K} \mid c\bigr).
\end{equation}
The release default, $K = 8$ and $w = 2.0$, is 16 generator passes per chunk of text.

\paragraph{Vocoder.} The vocoder follows the design of Vocos \citep{siuzdak2024vocos} and has no upsampling layers. Its 8 ConvNeXt blocks and a linear head predict a spectrum per mel frame, 1,025 log-magnitudes and 1,025 phases, and an inverse short-time Fourier transform (iSTFT) rebuilds the waveform from these overlapping spectra.

Training adds GAN losses, in which discriminator networks \citep{kong2020hifigan,jang2021univnet} learn to tell real audio from generated and the vocoder learns to fool them. The loss is $45\,\mathcal{L}_\text{mel} + \mathcal{L}_\text{adv} + 2\,\mathcal{L}_\text{feat}$, which combines L1 on log-mels at 4 resolutions, a least-squares adversarial term \citep{mao2017lsgan} and a match on the discriminators' internal activations.

\paragraph{The inference path.} Synthesis normalizes the text and splits it into chunks of at most 300 characters, about 19\,s at the release defaults. Every sentence of Freya-TR-Eval \citep{pamuk2026freyatts}, the benchmark we score on, is one chunk. Predicted durations are divided by a speed of 0.95, rounded and clamped to 1 to 60 frames (640\,ms). A doubling rule lengthens repeated sounds first (Section~\ref{sec:fixes}).

\begin{figure}[t]
\widefig{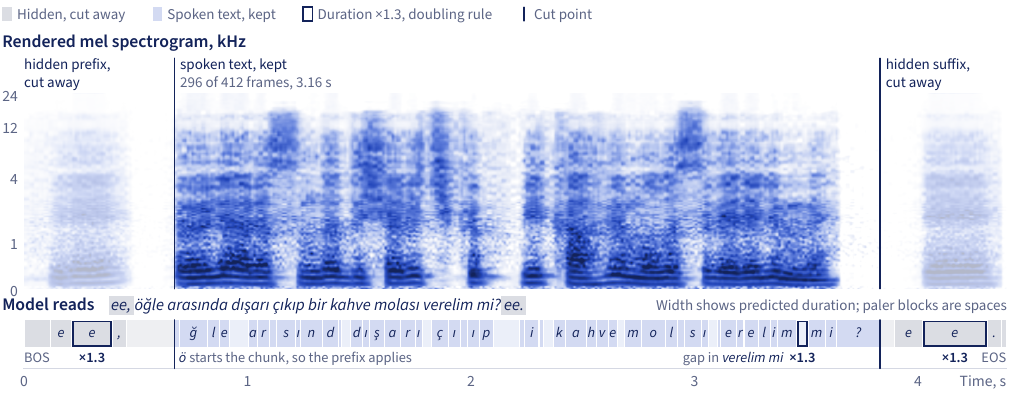}
\caption{\textbf{The model renders more text than the listener hears.} Version 1.0.1, seed 0. Some symbol blocks are too narrow for their letter; the ``Model reads'' row prints the full string. This Freya-TR-Eval sentence \protect\citep{pamuk2026freyatts,freya2026treval} uses the hidden prefix, the hidden suffix and the doubling rule. Across the benchmark's 495 sentences, 4 with the prefix, predicted durations discard 16.8\% of rendered frames. Longer chunks lose a smaller share, and this sentence loses more because it has the prefix.}\label{fig:inference}
\end{figure}

Every chunk gets a hidden \code{"~ee."} after it, meant to put the real last word mid-utterance. We have no measurement isolating a benefit at the final settings, and its WER effect went both ways in earlier seed-0 runs with the first vocoder fine-tune. A chunk that starts with \tr{ö} or \tr{ı} also gets a hidden \code{"ee,~"} before it, for the defect of Section~\ref{sec:fixes}. All networks run on the whole string, and the audio is cut at the predicted spans with 5\,ms fades (Figure~\ref{fig:inference}).

Between chunks the package adds 0.12\,s of silence after a sentence end and 0.06\,s otherwise. The same text, seed and settings give the same audio on one software stack.

\paragraph{Output chain.} Version 1.0.1 ends with fixed signal processing (Table~\ref{tab:parts}), the only default that differs from version 1.0.0. An equalizer (EQ), a gain that varies with frequency, uses 5 second-order filters \citep{bristowjohnson2021cookbook}. A de-esser turns down loud energy above 5.5\,kHz, the band of hissing sounds such as \textit{s}. A fixed +1.6\,dB gain offsets the mean loudness drop from those two, 1.596\,dB, measured with the BS.1770 standard \citep{itu2023bs1770} on 24 clips at 48\,kHz. A soft limiter guards the peaks, and 96 test clips over 4 output rates did not reach its threshold.

Version 1.0.0's EQ, the ``1.0.0 tone'' of our figures, raised the bass and cut the treble (Figure~\ref{fig:eq}a). We chose the 1.0.1 EQ by ear after one outside comment on that tone, and automatic scores moved slightly against it (Section~\ref{sec:eval}).

\paragraph{Where the compute goes.} At the defaults a 5.11\,s sentence costs 14.49\,GMAC ($10^9$ multiply-accumulates), or 2.84\,GMAC per second of output. The generator dominates because it runs 16 times per chunk and the vocoder once. Four steps with guidance cost 1.65\,GMAC per second, 8 steps without guidance 1.63 and 1 step without guidance 0.59. These counts do not measure time.

\section{Data from one teacher}\label{sec:data}
\standfirst{Unmodified VoxCPM2, cloning one synthetic clip, generated all our training audio on one 8-GPU machine. Automatic checks kept 1,325,846 of 1,361,734 clips, about 1,790\,h, but Whisper checked only 7,500 of the 444,437 short clips, so no rule tested whether the rest read their text correctly.}
\suppressfloats[t]

\paragraph{The teacher.} All training audio was generated by unmodified VoxCPM2 \citep{zhou2026voxcpm2} (\code{openbmb/VoxCPM2}, revision 32279eff, Apache-2.0), cloning the voice of one 9.88\,s reference clip that is itself a synthetic render. On all 1,361,734 clips it ran with classifier-free guidance 2.0 and 10 steps in ultimate mode, which gives the model the clip's transcript and two encodings of its audio. VoxCPM2's main checkpoint has about 2.29B parameters, about 2.38B with its AudioVAE (the part that decodes to 48\,kHz audio), roughly 300 to 313 times the student. Its authors report over 2 million hours of training speech in 30 languages, with no separate figure for Turkish.

\paragraph{The voice.} The training set contains no recordings of people: every clip was synthesized by VoxCPM2 from one synthetic reference clip. The voice is synthetic, chosen by listening from the renders of an earlier in-house model. That model and VoxCPM2 were themselves trained on recordings of real speakers, and the earlier model's voice-locking started from a reference recording of a consenting voice actor. We have not audited the voice's similarity to any real speaker. The model has no speaker input and cannot clone voices.

Voice-locking fixes one voice across renders. We clustered renders of that earlier model and, by listening, fixed the voice as 38 of them, with a median F0 (fundamental frequency, heard as pitch) of 130.4\,Hz and a 5th to 95th percentile band of 125.1--137.5\,Hz. VoxCPM2 then cloned the 24\,kHz render closest to the cluster center, and one 48\,kHz output became the reference clip for all training data, the chosen reference below.

\paragraph{The bake-off.} We compared 9 candidates, each a teacher with a reference clip, on 120 renders each, and pitch set the chosen reference apart (Table~\ref{tab:bakeoff}). A second refinement, which repeats the cloning step on the chosen reference, has a lower CER, and the earlier model's render a higher identity. Identity is the cosine similarity of a render's ECAPA speaker embedding \citep{desplanques2020ecapa} to the centroid of the 38 renders.

\begin{table}[t]
\caption{\textbf{The chosen reference put 83.3\% of renders in the pitch band; no other candidate passed 50.0\%.} Four of the 9 candidates, each with 2 renders of each of 60 prompts in 10 categories. CER comes from Whisper large-v3 at beam 5, and adversarial and foreign prompts inflate it for all of them. F0 is the mean of per-render medians. In band means 125--138\,Hz.}\label{tab:bakeoff}
\centering\tabfont
\begin{tabular}{@{}lrrrrr@{}}
\toprl
\thd{Teacher and reference} & \thd{CER, \%} & \thd{UTMOS} & \thd{Identity} & \thd{F0, Hz} & \thd{In band, \%} \\
\midrule
\us VoxCPM2, chosen reference & 2.95 & 3.387 & 0.697 & 129.42 & 83.3 \\
VoxCPM2, earlier model's render & 3.03 & 3.295 & 0.748 & 139.33 & 45.0 \\
VoxCPM2, second refinement & 2.82 & 3.365 & 0.631 & 125.13 & 50.0 \\
Alania-2 API, earlier model's render & 3.12 & 2.648 & 0.623 & 146.71 & 23.3 \\
\botrl
\end{tabular}
\end{table}

Alania-2, our closed production model, was only a candidate and produced no training audio. The table omits other VoxCPM2 settings and refined references, some of which scored higher on UTMOS or lower on CER than the chosen reference, with fewer renders in band. On Freya-TR-Eval (our own harness, 1 seed) the chosen reference gives WER 0.92\% and CER 0.200\%, and the second refinement 1.15\% and 0.289\%. That gap, on one seed, counted against the second refinement, so the benchmark had a part in this choice (Section~\ref{sec:limits}).

\begin{figure}[t]
\widefig{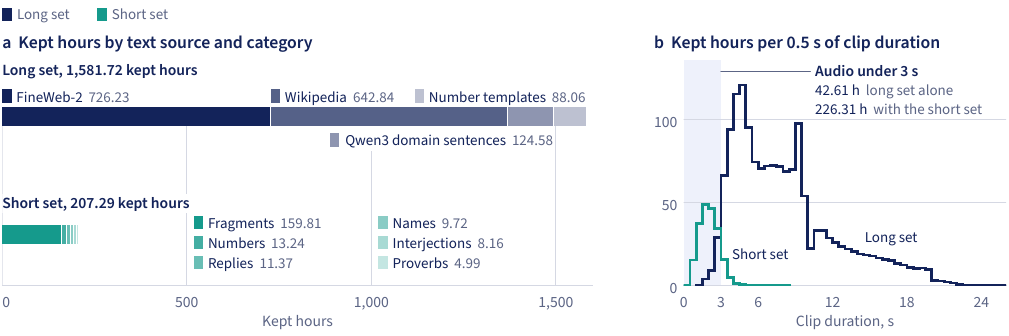}
\caption{\textbf{The short set lifts kept audio under 3\,s from 42.61\,h to 226.31\,h.} Neither validation nor the benchmark shows what the short-clip fine-tune did for short input (Section~\ref{sec:training}). Hours count unique finished audio, after edge-silence trim and before the \texttimes{}6 weighting of the short-clip fine-tune.}\label{fig:corpus}
\end{figure}

\paragraph{The long set.} We generated 917,297 lines, sentences up to about 25\,s, and kept 898,046 clips, 1,581.72\,h (Figure~\ref{fig:corpus}a), with a median of 5.22\,s. The text comes from FineWeb-2 \citep{penedo2025fineweb2}, Wikipedia, service-domain sentences written by Qwen3 \citep{yang2025qwen3} and our own number templates.

\paragraph{The short set.} We built it because the main run's word errors concentrated in sentences of 3 to 4 words and little of the long set was under 3\,s (Figure~\ref{fig:corpus}b). That length effect is confounded with register, the style of the sentence (Section~\ref{sec:eval}). We kept 427,800 of 444,437 clips, 207.29\,h, with a median of 1.64\,s. Fragments of 2 to 6 words cut from the long-set text pools make up 159.81\,h. On the shipped path only the short-clip fine-tune used this set, weighted \texttimes{}6, and the vocoder never saw it.

A bug reached this set. Turkish pairs \tr{i} with \tr{İ} and \tr{ı} with \tr{I}, so when Python's \code{str.upper} turned a leading \tr{i} into ASCII \tr{I}, the normalizer lowered it to \tr{ı}. As a result 4,580 of the 5,589 kept \tr{ı}-initial lines are misspelled (for example \tr{ıyi}, \tr{ıki}). They stayed in the short set, and we have no measurement isolating their effect on the model.

\paragraph{Keeping the benchmark out.} We excluded benchmark sentences from the training text, which dropped 11 long-set lines and 5,562 short-set lines. A later recheck of the final text with the same guard found no exact, contained or near match (rapidfuzz string similarity of 85 or more), though it did not re-run the short set's rule for near matches to part of a sentence.

\begin{figure}[t]
\widefig{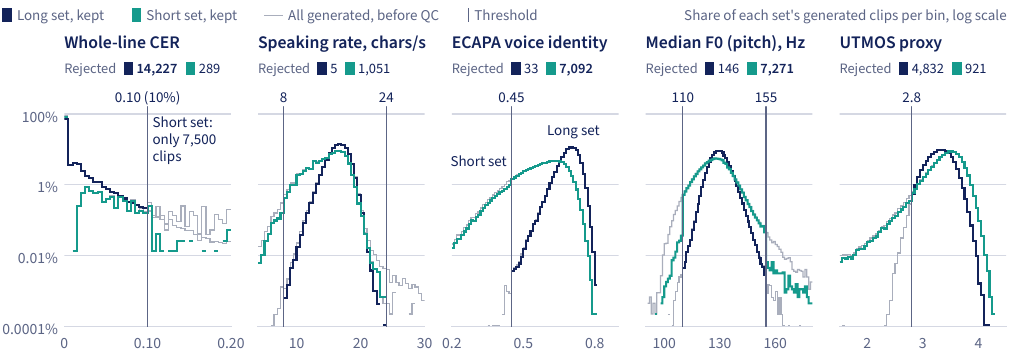}
\caption{\textbf{QC kept 97.4\% of 1,361,734 generated clips, and the kept short set sits further from the voice than the kept long set.} Mean identity is 0.582 against 0.694, and 57.0\% of short clips fall in 125--138\,Hz against 82.4\% of long-set clips. Short-set clips past a threshold passed under relaxed rules. Axes clip outliers.}\label{fig:qc}
\end{figure}

\paragraph{Quality control.} A quality-control (QC) worker rejected a clip at the first rule it failed (Table~\ref{tab:qc}, Figure~\ref{fig:qc}).

\begin{table}[t]
\caption{\textbf{CER rejected the most long-set clips, and F0 and identity the most short-set clips.} QC applied the rules in this order.}\label{tab:qc}
\centering\tabfont
\begin{tabular}{@{}l>{\raggedright\arraybackslash}p{34mm}>{\raggedright\arraybackslash}p{42mm}rr@{}}
\toprl
\thd{Rule} & \leavevmode\thd{Rejects when} & \leavevmode\thd{Relaxed in the short set} & \thd{Long set} & \thd{Short set} \\
\midrule
Length limits & Clip too long or too short & & 8 & 13 \\
CER & Above 10\% & Up to 4 words: passes if the clip's text appears in the transcript. Interjections: off & 14,227 & 289 \\
Speaking rate & Outside 8--24 characters per second & Up to 2 words: no minimum & 5 & 1,051 \\
ECAPA identity & Below 0.45 & Under 1.2\,s: off & 33 & 7,092 \\
Median F0 & Outside 110--155\,Hz & Interjections: off & 146 & 7,271 \\
UTMOS & Below 2.8 & Up to 4 words: off & 4,832 & 921 \\
\midrule
All rejections & & & 19,251 & 16,637 \\
\botrl
\end{tabular}
\end{table}

\paragraph{What Whisper checked.} The Whisper speech recognizer \citep{radford2022whisper}, with the large-v3 checkpoint in faster-whisper \citep{fasterwhisper}, checked all 917,297 long-set clips. We turned it off for the short set after 15 shards, 7,500 of its 444,437 clips, because it pads each clip to 30\,s, which made QC about 10 times slower than generation. The other 436,937 clips have empty transcripts and a placeholder CER of 0. Speaking rate, ECAPA identity, F0 band and UTMOS rules still applied to them, relaxed for very short ones.

In the long set the 10\% CER gate reads the whole line, and 82\% of \tr{ö}-initial lines with a wrong first word passed it (Section~\ref{sec:fixes}). Clips Whisper could not transcribe never reached long-set training, and the same checkpoint scores the benchmark, which may flatter WER (Section~\ref{sec:limits}).

\begin{figure}[t]
\widefig{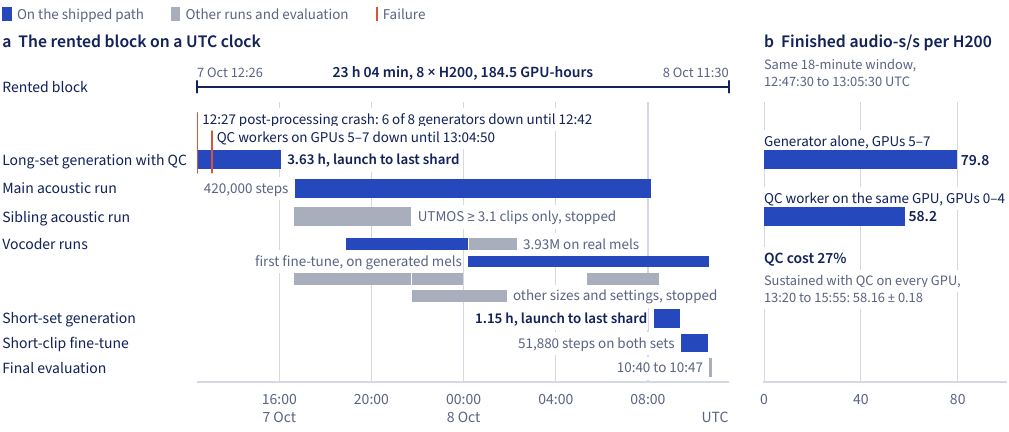}
\caption{\textbf{Generation itself took 3.63\,h (long set) and 1.15\,h (short set) from launch to last shard; 23\,h 04\,min is the whole block.} Cobalt marks the shipped path, teacher runs included; the final vocoder fine-tune ran after the block.}\label{fig:block}
\end{figure}

\paragraph{Generation.} We used one AWS p5en.48xlarge with 8 NVIDIA H200 GPUs, rented for 23\,h 04\,min (the block of Figure~\ref{fig:block}a). Soon after launch, segfaults in Numba's shared cache crashed the post-processing pools of 6 of 8 generators (one VoxCPM2 process per GPU). A private cache per worker and a self-rebuilding pool fixed this.

A Whisper-based QC worker on the same GPU as the generator cost 27\% of throughput (Figure~\ref{fig:block}b). From 12:47:30 to 13:05:30, with the QC workers on GPUs 5--7 down until 13:04:50, those GPUs finished 79.8 seconds of audio per second (audio-s/s) per H200, and GPUs 0--4, each with a QC worker, finished 58.2. This is one unplanned comparison, in an 18-minute window on different GPUs of one machine. Sustained with QC on every GPU, the rate was 58.16 $\pm$ 0.18.

\begin{lesson}\textbf{Lesson.} Count QC in the generation plan. Decide before the rental which clips must be transcribed, and leave GPU time for them.\end{lesson}

\section{Training the acoustic model}\label{sec:training}
\standfirst{The bf16 autocast weight cache kept gradients from the generator for the first 6,000 of the main run's 420,000 steps. The short-clip fine-tune in the shipped weights is not distinguishable from the main run on the benchmark, and the shipped lineage took 114.5 GPU-hours.}

The shipped acoustic weights are an exponential moving average (decay up to 0.9995), taken after 420,000 steps of the main run on the long set and 51,880 steps of the short-clip fine-tune on both sets. Both use AdamW (betas 0.9 and 0.98, weight decay 0.01), gradient clipping at 1.0 and bf16 autocast (mixed precision).

The main run used a base learning rate of $3\times10^{-4}$ and 3 H200 GPUs, then 2 from step 55,000, with 96k mel frames per GPU from step 2,000. In parallel we trained a sibling run on long-set clips with a QC UTMOS of at least 3.1.

\paragraph{A silent bug.} We computed shortcut targets under \code{no\_grad} inside the same autocast region as the training pass. Autocast caches the weights it casts, so the training pass received detached copies and the generator got no gradients. Setting \code{cache\_enabled=False} fixed it.

\begin{figure}[t]
\widefig{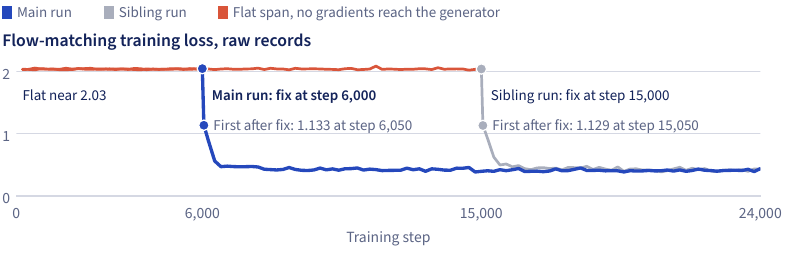}
\caption{\textbf{The bf16 autocast weight cache gave the generator no gradients for the first 6,000 steps of the main run and 15,000 of the sibling run.} Validation loss on the moving-average weights stayed at exactly 2.0359 in the main run and 2.0342 in the sibling run, and at step 13,000 of the sibling run the generator's output weights still had zero norm.}\label{fig:bug}
\end{figure}

The signature is a flow-matching loss flat near 2.03 in training and exactly constant in validation on the moving-average weights, with a shortcut loss of 0 (Figure~\ref{fig:bug}). The encoder, prior and duration predictor trained normally (prior loss 0.529 to 0.039 by step 6,000), so total loss still fell.

\begin{lesson}\textbf{Lesson.} After the first step, assert that every trainable module has a gradient. Our first launch crashed on an unused-parameter error, which was this bug's symptom. We silenced it with \code{find\_unused\_parameters=True} before we knew the cause.\end{lesson}

\paragraph{The short-clip fine-tune.} It starts from step 420,000 of the main run with a fresh AdamW and a learning rate of $10^{-4}$, and spends its last 23,880 steps at a floor of $5\times10^{-6}$. With the short set repeated six times, the two sets total 2,802.62\,h, of which short clips are 44.07\%. Validation runs on the first 256 clips of the validation split, all from the long set, and its flow-matching loss ended at 0.3969, where it stood at step 2,000. A flat validation loss was also the bug's signature, so we compared checkpoints. The generator's moving-average weights moved by 6.8\% in relative L2 norm between the main run's last step and the shipped file.

On the benchmark the short-clip fine-tune is not distinguishable from the main run. Its WER minus the main run's is $-$0.08 [$-$0.32,~+0.15] at 4 sampling steps and $-$0.05 [$-$0.25,~+0.15] at 8 (95\% intervals from a paired bootstrap over sentences, 3 seeds, our own harness, no output chain, same vocoder). Validation holds no short clips and these whole-benchmark intervals include 0, so we cannot say what the fine-tune did for short input.

The sibling run's filter has no fair test either. In the second evaluation round during training, with the same vocoder run, it had WER 6.28\% and UTMOS 1.78 against 5.21\% and 2.26 for the main run (seed 0, our own harness), at different learning rates and step counts.

\paragraph{Compute.} MAS took 61.2\% of main-thread profiler samples in an acoustic step. Sending only its log-likelihoods to the CPU and giving it more threads raised median throughput on 3 H200s from 15,954 to 20,628 audio-s/s (+29\%).

The shipped lineage took 114.5 GPU-hours, 36.4 for the acoustic model on H200s and 78.0 for the vocoder (56.0 on H200s, 22.0 on A100s and an L40S). All training runs took 159.0, of which 130.6 ran on the machine of Section~\ref{sec:data}, where setup, generation and QC took another 33.7. Evaluation is not counted.

\section{The vocoder kept WER high}\label{sec:vocoder}
\standfirst{On our own harness, at adjacent pre-release acoustic checkpoints with the same seed, fine-tuning the vocoder on the acoustic model's own mels cut WER from a mean of 3.89\% to 1.75\%, while a vocoder frozen over the same stretch stayed at 3.63--4.14\%.}
\suppressfloats[t]

\begin{figure}[t]
\widefig{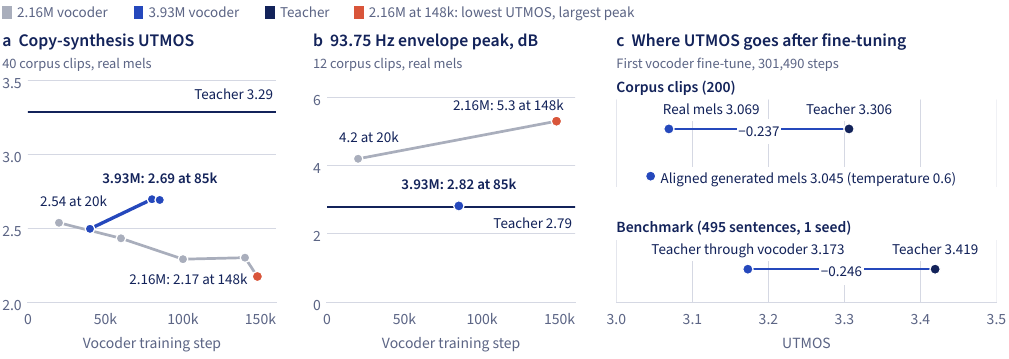}
\caption{\textbf{A 2.16M vocoder got worse with training on real mels; after fine-tuning, the 3.93M vocoder alone costs about 0.24 UTMOS.} Pre-release checkpoints. (a) Checkpoints differ by about $\pm$0.1 UTMOS. The 2.16M vocoder's mel loss read 0.3927, 0.4225, 0.3544 and 0.3809 at steps 20k, 60k, 100k and 148k. (b) The peak of the amplitude envelope's spectrum at the 93.75\,Hz frame rate, over its median from 60 to 130\,Hz. (c) Benchmark renders on our own harness. ``Real mels'' and ``Teacher through vocoder'' are the same measurement on different clips. Aligned generated mels use each clip's MAS durations.}\label{fig:vocoder}
\end{figure}

Our 2.16M vocoder got worse as it trained on real mels. Its copy-synthesis UTMOS on 40 corpus clips, meaning clips from the training set, fell from 2.54 at step 20,000 to 2.17 at step 148,000, against 3.29 for the teacher (Figure~\ref{fig:vocoder}a). Mel loss did not track the fall.

Its output also grew a ripple at 93.75\,Hz, the vocoder's frame rate. The envelope peak at that rate rose from 4.2\,dB at step 20,000 to 5.3\,dB at step 148,000, against 2.79\,dB for the teacher (Figure~\ref{fig:vocoder}b). The larger 3.93M vocoder of the shipped lineage reached UTMOS 2.69 with a peak of 2.82\,dB at step 85,000.

\paragraph{Fine-tuning on generated mels.} This real-mel vocoder still left WER above 3.6\%. It had trained only on real mels, and at inference it receives the generator's mels, so we fine-tuned a copy of its step 157,000 on those, the first vocoder fine-tune.

\begin{figure}[t]
\widefig{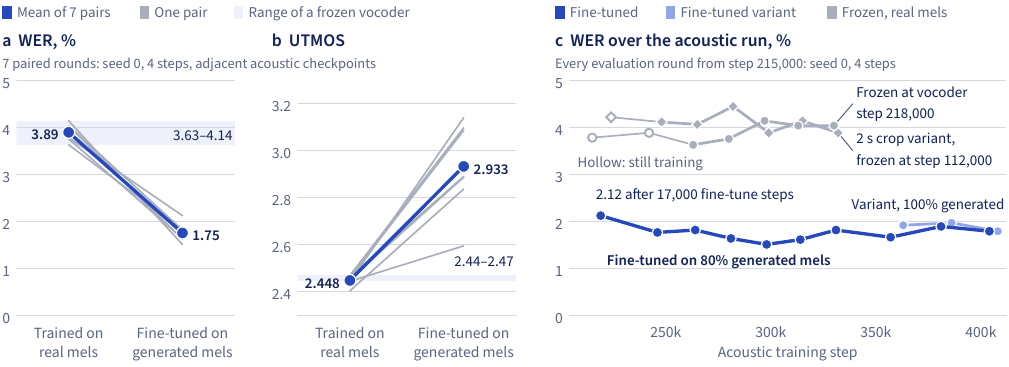}
\caption{\textbf{At adjacent acoustic checkpoints, seed 0 and 4 sampling steps, fine-tuning the vocoder on generated mels cut WER from a mean of 3.89\% to 1.75\%.} (a, b) Every paired 95\% interval of the WER drop excludes 0. UTMOS after fine-tuning rises with fine-tune steps, from 2.595 at 17,000 to 3.140 at 135,000. (c) The 2\,s crop variant, a real-mel vocoder trained on 2\,s crops and frozen at step 112,000, held WER at 3.88--4.44\% and UTMOS at 2.40--2.42. The 100\% variant, a fine-tune with generated mels on every crop, ran at unequal steps, so we draw no conclusion from it. Benchmark, our own harness, no output chain, pre-release checkpoints.}\label{fig:gta}
\end{figure}

The evaluation loop during training gave 7 pairs. Each scores the real-mel vocoder on one acoustic checkpoint and the first vocoder fine-tune on an adjacent one, at seed 0 and 4 sampling steps. Over these pairs the fine-tune cut WER from a mean of 3.89\% to 1.75\% (Figure~\ref{fig:gta}a). The drop per pair runs from 1.66 to 2.63 points, and every paired sentence-bootstrap 95\% interval excludes 0. CER fell from 0.984\% to 0.385\% and UTMOS rose from 2.45 to 2.93 (Figure~\ref{fig:gta}b).

A pair's two checkpoints sit 1,000 to 4,000 steps apart. Further acoustic training does not explain the drop, because the real-mel vocoder, frozen at vocoder step 218,000, stayed at WER 3.63--4.14\% and UTMOS 2.44--2.47 while the main run moved from acoustic step 263,000 to 330,000, the last 5 of the 7 pairs (Figure~\ref{fig:gta}c). It was still training for the first two pairs.

For this fine-tune the acoustic model samples one fixed mel per clip, at 4 steps and classifier-free guidance 2.0 on the clip's own MAS durations, so frame counts match the teacher's audio, which stays the target. Each training crop takes the generated mel with probability 0.8 and the real mel otherwise, which is the ``80\%'' of Figure~\ref{fig:gta}c. The shipped vocoder's stages use AdamW (betas 0.8 and 0.99, weight decay 0.01, clipping at 10.0) on 1.003\,s crops, 32 per GPU, with no moving average.

These are pre-release checkpoints at temperature 1.0 with no output chain, on our own harness. We never scored a real-mel vocoder at the final acoustic checkpoint. WER here is Whisper's, and no human listening test of this model exists. With more training the main run and the first vocoder fine-tune reached 1.38\% on 3 seeds (acoustic step 420,000, fine-tune step 301,490, 4 sampling steps), a later result from outside the 7 pairs.

\begin{lesson}\textbf{Lesson.} Score the whole chain during training, and fine-tune the vocoder on generated mels early. Our first pair came after 17,000 fine-tune steps, about 34 minutes on one H200, and moved WER from 3.78\% to 2.12\%.\end{lesson}

\paragraph{Precedent.} Tacotron 2 and HiFi-GAN \citep{shen2018tacotron2,kong2020hifigan} also report gains from vocoders trained on predicted features. Those are human ratings on teacher-forced features, where the acoustic model sees the real previous frames, so the effect sizes do not compare with ours (Section~\ref{sec:related}).

\paragraph{What remains.} After fine-tuning, the vocoder alone costs about 0.24 UTMOS (Figure~\ref{fig:vocoder}c). Copy synthesis through the first vocoder fine-tune at step 301,490, the parent of the shipped vocoder, scores 3.069 against 3.306 for the teacher on 200 corpus clips, and aligned generated mels at temperature 0.6 score 3.045. On the teacher's 495 benchmark renders it scores 3.173 against 3.419, with WER 1.251\% against 0.919\% (1 seed, our own harness, unseen text).

The last 76,860 of the shipped vocoder's updates, the final vocoder fine-tune on mels from the short-clip fine-tune, gave no demonstrated gain. On 3 seeds through the release API on our own harness, before the fixes of Section~\ref{sec:fixes}, WER moved from 1.115\% to 1.089\% (131 to 128 word errors) and UTMOS from 3.268 to 3.256. The final vocoder fine-tune saw mels at 4 steps, temperature 0.8 and MAS durations, while the release samples at 8 steps, temperature 0.6 and predicted durations. The vocoder never saw short-set audio.

\section{The teacher's ö defect and two reading fixes}\label{sec:fixes}
\standfirst{By Whisper, the teacher gets first words right on only 26\% of \tr{ö}-initial kept long-set lines, and the student learned this. A hidden spoken prefix hides the defect without removing it, with 12 of 12 first words right on 4 benchmark sentences. The paired interval for a duration rule on doubled sounds includes 0.}
\suppressfloats[t]

\begin{figure}[t]
\widefig{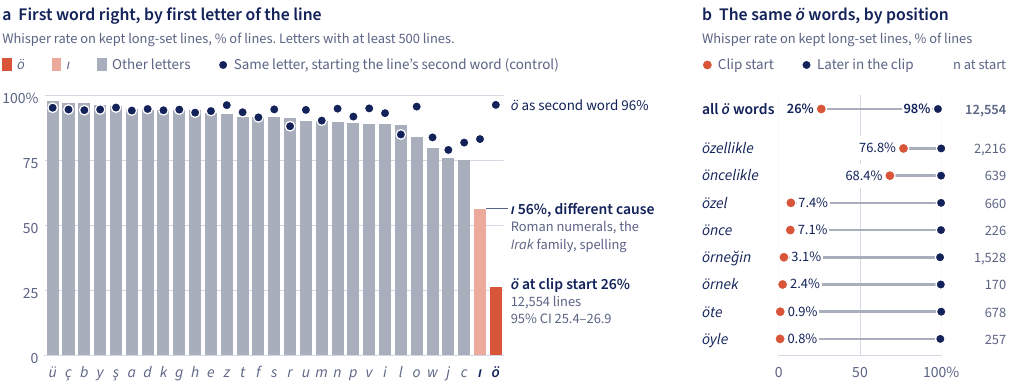}
\caption{\textbf{By Whisper, only 26\% of \tr{ö}-initial kept long-set lines have a right first word, against 94\% for other capital first letters.} (a) \tr{ı} errors have other causes: the word \tr{ı} from Roman numerals, \tr{Irak} and its inflections, heard as \tr{drak}, and spelling. (b) Later positions are pooled and use aligned transcripts.}\label{fig:onset-letters}
\end{figure}

\paragraph{The defect in the teacher.} Only 26\% of the 12,554 kept long-set lines whose first word begins with the vowel \tr{ö}, capital or not, have that word right in their QC Whisper transcripts (95\% CI \mbox{25.4--26.9}; Figure~\ref{fig:onset-letters}a). Lines with other capital first letters reach 94\%, and capital \tr{Ö} alone, the like-for-like figure, scores 23.66\%. No listening test of teacher clips exists.

The same \tr{ö} words score about 98\% later in the clip (Figure~\ref{fig:onset-letters}b), which ties the defect to the clip start. The clip-start rate depends on the word, with 76.8\% for \tr{özellikle} against 0.8\% for \tr{öyle}. Wrong transcripts often open with a phantom syllable, as in \tr{latte} for \tr{önce}.

We did not establish the cause inside the teacher. With 4 renders per configuration on \tr{ö} prompts, Whisper read 4 of 4 first words right in reference-only mode (an encoding of the reference clip's audio, no transcript), and in ultimate mode 0 or 1 of 4 with most references and 4 of 4 with one. Samples this small are only suggestive.

\begin{figure}[t]
\widefig{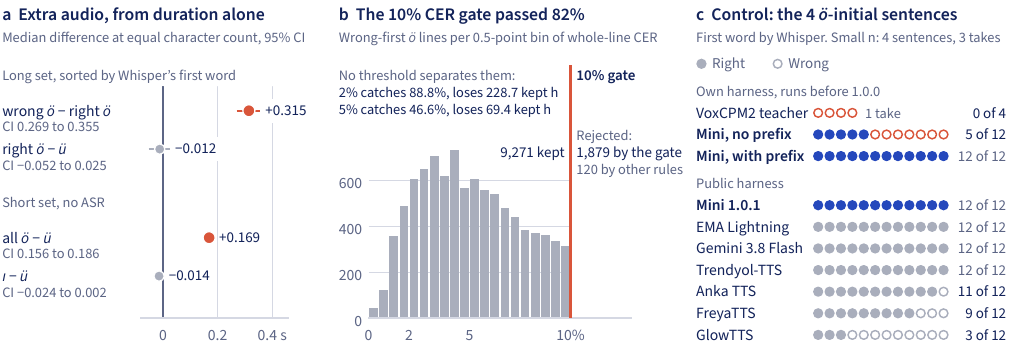}
\caption{\textbf{The \tr{ö} defect adds audio measurable without speech recognition (ASR), 82\% of \tr{ö} lines with a wrong first word passed the 10\% CER gate, and most other systems' first words transcribe correctly on the same sentences.} (b) Rejected long-set lines are counted, not binned. (c) Teacher with the chosen reference. Small n: 4 sentences.}\label{fig:onset-evidence}
\end{figure}

\paragraph{How we know it is audio.} All of this rests on Whisper, which could be misreading good audio. Two checks point to the audio instead. At equal character count, long-set \tr{ö} lines that Whisper reads with a wrong first word run a median 0.315\,s longer than those with a right one, which suggests extra sound, and right \tr{ö} lines match lines starting with \tr{ü}, the highest-scoring letter, within noise (Figure~\ref{fig:onset-evidence}a). The short-set rows use no recognizer, and there all \tr{ö} lines run 0.169\,s longer than \tr{ü} lines while \tr{ı} lines do not stand out.

Whisper also does not invent the syllable. On the 4 \tr{ö}-initial benchmark sentences, a small control, it reads 11 or 12 of 12 first words (3 takes each) from most other systems on the public harness, including EMA Lightning, whose card reports the same teacher model with another voice. On our own harness it reads 0 of 4 from the teacher with the chosen reference, in 1 take (Figure~\ref{fig:onset-evidence}c).

\paragraph{How the student inherited it.} Our 10\% whole-line CER gate missed most of the defect. Of the 11,270 generated long-set lines with a wrong \tr{ö} first word, the gate rejected 1,879, other rules rejected 120, and 9,271 (82\%) passed (Figure~\ref{fig:onset-evidence}b). The lines that passed stayed in the training data, and the student learned the defect. Without the prefix it gets 5 of 12 first words right on the 4 benchmark sentences (our own harness, 3 seeds, before the final vocoder fine-tune). Informally, one of us heard student renders of three \tr{ö}-initial sentences as broken.

\begin{lesson}\textbf{Lesson.} Check clip starts separately from whole-line CER. A wrong first word costs a median of only 5 characters on a median 91-character line, so 82\% of \tr{ö} lines with a wrong first word passed our 10\% gate.\end{lesson}

\begin{figure}[t]
\widefig{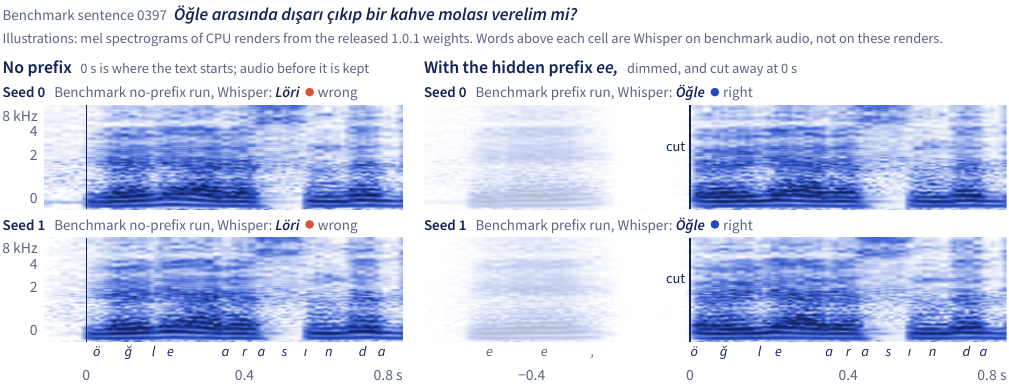}
\caption{\textbf{On benchmark sentence 0397, Whisper wrote \tr{Löri} in the no-prefix run and \tr{Öğle} in the prefix run, at seeds 0 and 1.} Read the words above the cells; the spectrograms only show where the prefix sits and where we cut it. The transcripts come from two own-harness runs before 1.0.0, and the no-prefix run predates the final vocoder fine-tune, so the figure does not isolate the prefix. Sentence: Freya-TR-Eval \protect\citep{pamuk2026freyatts,freya2026treval}.}\label{fig:onset-spec}
\end{figure}

\paragraph{The hidden prefix.} The defect sits at the clip start, so we make the model say something else first and cut that audio away. A chunk that starts with \tr{ö} or \tr{ı} gets a hidden \code{"ee,~"} in front of it, cut at its predicted span (Section~\ref{sec:model}). With the prefix, Whisper reads 12 of 12 first words on the 4 sentences on both harnesses. Figure~\ref{fig:onset-spec} shows one of them. With the final vocoder fine-tune, per-clip no-prefix data exists only for seed 0, where Whisper again writes \tr{Löri} for \tr{Öğle}.

In own-harness runs before 1.0.0, with the final vocoder fine-tune, the prefix lowers 3-seed WER from 1.089\% to 1.013\% and CER from about 0.25\% to 0.204\%. On seed 0 the gain comes from 2 of the 4 sentences, where we had also seen the symptom. The missing per-clip data leaves this step with no 3-seed paired interval.

Applied to every chunk in a probe of 84 new sentences, the prefix broke other words, such as \tr{Jöle}, heard as \tr{göle} or \tr{Göle} on 2 of 3 seeds. So we restrict it to \tr{ö} and \tr{ı}. The \tr{ı} part rests on that probe alone, because the benchmark has no \tr{ı}-initial sentence. In the probe, first words starting with capital \tr{I} went from 6 to 8 of 9 right, and \tr{Ilgaz} was heard as \tr{Bulgaz}, \tr{Dılgaz} or \tr{Rılgaz} without the prefix. The teacher's \tr{ı} errors have other causes (Figure~\ref{fig:onset-letters}a).

\paragraph{Doubled sounds.} The doubling rule multiplies by 1.3 the predicted duration of the second of two identical symbols, or of the word gap between them (the predicted span of the space symbol). Such pairs occur in 167 of the benchmark's 495 sentences (112 pairs within words, 111 across word gaps).

\begin{figure}[t]
\widefig{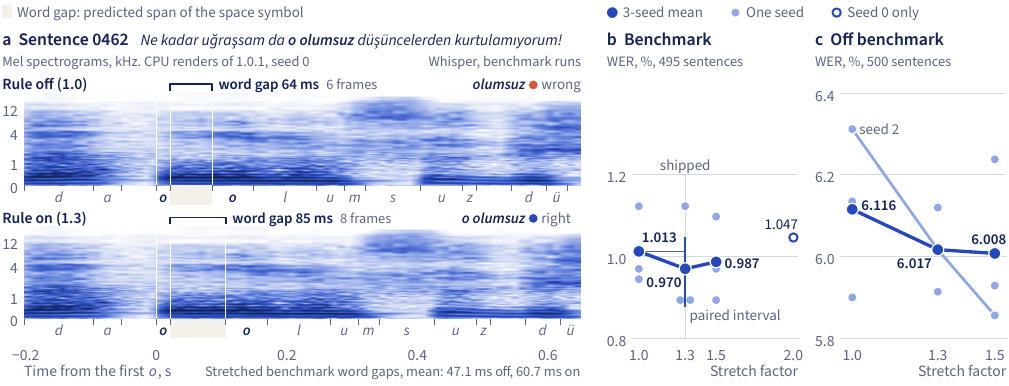}
\caption{\textbf{The doubling rule moves 3-seed WER from 1.013\% to 0.970\%, with a paired interval that includes 0; it fixes \tr{o olumsuz} on 3 of 3 seeds on both harnesses, and the doubled \tr{rr} in \tr{berraktı} stays wrong.} Own-harness runs use 1.0.0 settings and public takes 1.0.1. (b) The interval is measured from the mean at stretch factor 1.0. (c) We computed no intervals, and nearly all the gain comes from seed 2. WER is near 6\% because of digits and foreign names the scorer does not match. Sentence: Freya-TR-Eval \protect\citep{pamuk2026freyatts,freya2026treval}.}\label{fig:gem}
\end{figure}

The rule targets doubled sounds heard as one, as in \tr{o olumsuz} read as \tr{olumsuz}, a pattern we found in benchmark errors. Before the rule, 21 sentences failed on all 3 seeds of our own harness. EMA Lightning also fails 13 of them on all 3 of its seeds and gets 6 right on all 3, so Whisper can read those 6 from another model's audio. Of those 6, 3 show the pattern (Figure~\ref{fig:gem}a).

A result on Turkish consonants motivates the rule. \citet{lahiri1988timing} find that doubled Turkish [t] and [k] between vowels, from 3 male speakers, are held almost three times as long as single ones, about 120\,ms longer. Vowels, other consonants and word-gap pairs, which the rule also lengthens, fall outside their data. On the benchmark we tried stretch factors of 1.3, 1.5 and, on seed 0 only, 2.0, and we shipped 1.3 (Figure~\ref{fig:gem}b).

Over 3 seeds on our own harness, the rule moves WER from 1.013\% to 0.970\% and CER from 0.204\% to 0.193\%. Both paired intervals include 0 (WER $-$0.043 [$-$0.135, +0.035], CER $-$0.011 [$-$0.035, +0.010] points). Word errors fall from 53 to 46 in sentences with such a pair and go from 66 to 68 in the rest. The rule came from these sentences, so these numbers are optimistic by selection.

Off the benchmark the direction agrees, and the change is not separable from seed noise. On 500 other sentences, 464 of them not in the training text by exact match (near-duplicates unchecked), 3-seed WER moves from 6.12\% to 6.02\% and CER from 1.53\% to 1.48\% (Figure~\ref{fig:gem}c). Nearly all the WER gain is one seed. For these runs and the stretch factors 1.5 and 2.0, we kept only the lines the scorer printed.

\paragraph{What did not help.} On our own harness, stretching every word gap by a factor of 1.25 gave 3-seed WER 1.055\% against 1.013\% without it.

Filtered training gave no demonstrated gain on the same harness. We fine-tuned the shipped acoustic model twice on filtered data and scored both runs with the shipped vocoder, not fine-tuned again. One run dropped lines with a bad start, gave good \tr{ö} and \tr{ı} starts more weight, and scored WER 1.174\% and CER 0.256\% on seed 0 without the prefix, against 1.047\% and 0.241\%. The other also dropped every long-set line with any character error and scored 1.106\% and 0.214\% over 3 seeds with the prefix, against 1.013\% and 0.204\%. We made no fix at the source, and regenerating the affected clips remains open (Section~\ref{sec:next}).

\section{Evaluation}\label{sec:eval}
\standfirst{On the 495 sentences of Freya-TR-Eval, on which we tuned settings, Mini 1.0.1 has WER 1.09\% and CER 0.220\% on the public harness. A paired bootstrap cannot distinguish these from EMA Lightning 1.0.1 or, on our own harness, from the teacher (1 seed), and no human listening test of Mini exists. EMA Lightning is ahead on full-band UTMOS and, on our own harness, Mini in the 8\,kHz band the benchmark's card asks for, while 16 sentences fail repeatedly for Mini and not for EMA Lightning.}

\paragraph{The benchmark.} Freya-TR-Eval \citep{pamuk2026freyatts} holds 495 Turkish sentences of 3 to 13 words in two registers: 406 everyday sentences written by an LLM and 89 short-native sentences from the public corpora Common Voice and CoVoST2. No reference contains a digit, so these scores do not test number readings. Three takes add up to 11,751 scored words and 80,922 characters. We report the benchmark card's secondary metric, band-matched WER and CER. We did not compute its primary metric, semantic WER, in which an LLM judge counts only errors that change the meaning.

\paragraph{Two harnesses.} The public harness \citep{patientdesk2026rerun} carries the headline numbers and all 15 systems, and a reader can re-run its open-weights rows (Appendix~\ref{app:repro}). Development runs, the teacher and 8\,kHz UTMOS are on our own harness, whose code is private. Both pass raw text to each system and score alike: 8\,kHz band matching, the faster-whisper large-v3 recognizer \citep{fasterwhisper} at beam 5, text normalization for scoring with identical output on all 3,772 strings we compared, corpus-level WER and CER per take from jiwer \citep{jiwer}, and UTMOS22-strong \citep{saeki2022utmos} from SpeechMOS v1.2.0 at 16\,kHz.

\paragraph{Results.} Mini 1.0.1's WER of 1.09\% on the public harness (Table~\ref{tab:results}) is second of the 15 systems we ran, behind EMA Lightning by 0.05 points, a gap the paired intervals below cannot distinguish from zero. Its CER of 0.220\% is the lowest of the 11 open-weights systems (0.225\% for EMA Lightning, 0.229\% for Trendyol-TTS) and the second lowest of all 15, behind our own closed Alania-2 voice Öncel (0.217\%). These gaps are 2 to 7 character errors in 80,922, and none is distinguishable from zero. EMA Lightning is lower on our own harness (0.198\% against 0.205\%). Mini ranks 11th of 15 at 3.18 on full-band UTMOS, below most large systems (Trendyol-TTS 3.82, Alania-2 F1 of 7 Oct 3.67). Ranks count Mini 1.0.1 only and Alania-2 at its top voice in each column. We tuned Mini's settings on these sentences, so final-settings numbers are optimistic by selection, and the scoring Whisper checkpoint also filtered the training data (Section~\ref{sec:limits}).

\begin{table}[t]
\caption{\textbf{On the public harness no WER or CER difference between Mini 1.0.1 and the rows from EMA Lightning down to Gemini 3.8 Flash-Lite is distinguishable.} No reference contains a digit. Rows are sorted by WER within each block, with Mini as two releases of the same weights, which differ only in the output chain (Figure~\ref{fig:eq}), and Alania-2 as three voices with their run dates. Alania-1, Alania-2 and Antalia-1 are also ours.}\label{tab:results}
\centering\tabfont\renewcommand{\arraystretch}{1.08}
\begin{tabular}{@{}llrrrrr<{\hspace{7pt}}@{}}
\toprl
\thd{System} & \thd{Weights} & \thd{WER, \%} & \thd{CER, \%} & \thd{Clean, \%} & \thd{UTMOS} & \thd{Parameters} \\
\midrule
\multicolumn{7}{@{}l}{\textbf{Public harness, 495 sentences, 3 takes}} \\
EMA Lightning 1.0.1 & open & 1.04 & 0.225 & 93.7 & 3.30 & 8.6M \\
Antalia-2 Mini 1.0.0, previous release & open & 1.05 & 0.200 & 93.1 & 3.25 & 7.62M \\
\us Antalia-2 Mini 1.0.1 & open & 1.09 & 0.220 & 92.9 & 3.18 & 7.62M \\
Alania-2 API, voice Öncel (12 Oct) & closed & 1.15 & 0.217 & 93.0 & 3.52 & 488M\rlap{\textsuperscript{a}} \\
Trendyol-TTS & open & 1.16 & 0.229 & 93.1 & 3.82 & 2.38B\rlap{\textsuperscript{d}} \\
Gemini 3.8 Flash TTS & closed & 1.23 & 0.289 & 92.9 & 3.53 & -- \\
Alania-2 API, voice Lina (12 Oct) & closed & 1.25 & 0.258 & 93.1 & 3.66 & 488M\rlap{\textsuperscript{a}} \\
Gemini 3.8 Flash-Lite TTS & closed & 1.34 & 0.273 & 91.9 & 3.63 & -- \\
Anka TTS & open & 1.42 & 0.302 & 91.6 & 3.52 & 336M\rlap{\textsuperscript{a}} \\
Alania-2 API, voice F1 (7 Oct) & closed & 1.44 & 0.281 & 91.7 & 3.67 & 488M\rlap{\textsuperscript{a}} \\
Chatterbox Multilingual & open & 1.91 & 0.429 & 88.8 & 3.51 & 801M\rlap{\textsuperscript{c}} \\
Piper, voice dfki & open & 3.41 & 0.733 & 79.4 & 3.71 & 16M \\
XTTS-v2 & open & 3.67 & 1.498 & 77.1 & 2.97 & 470M \\
Alania-1 API & closed & 5.99 & 1.809 & 67.3 & 3.19 & 3.3B\rlap{\textsuperscript{b}} \\
Antalia-1 & open & 6.28 & 2.489 & 66.7 & 2.72 & 305M\rlap{\textsuperscript{a}} \\
MMS-TTS & open & 6.38 & 1.480 & 65.7 & 3.79 & 36M \\
Coqui GlowTTS, lowercased input & open & 10.93 & 2.958 & 46.1 & 3.09 & 28M\rlap{\textsuperscript{a}} \\
FreyaTTS-small & open & 12.02 & 4.946 & 47.1 & 2.81 & 183M\rlap{\textsuperscript{a}} \\
\midrule
\multicolumn{7}{@{}l}{\textbf{Own harness, 495 sentences, 3 seeds (teacher 1 seed)}} \\
VoxCPM2, the teacher & open & 0.92 & 0.200 & 94.5 & 3.42 & 2.38B\rlap{\textsuperscript{d}} \\
Antalia-2 Mini 1.0.0, previous release & open & 0.97 & 0.193 & 93.7 & 3.25 & 7.62M \\
EMA Lightning 1.0.1 & open & 1.00 & 0.198 & 93.8 & 3.30 & 8.6M \\
\us Antalia-2 Mini 1.0.1 & open & 1.02 & 0.205 & 93.3 & 3.18 & 7.62M \\
\midrule
\multicolumn{7}{@{}l}{\textbf{Quoted from EMA Lightning's card, not run by us}} \\
ElevenLabs v4 & closed & 1.43 & 0.27\phantom{0} & -- & 3.30 & -- \\
ElevenLabs v4 Turbo & closed & 1.54 & 0.29\phantom{0} & -- & -- & -- \\
\botrl
\end{tabular}
\par\vspace{5pt}\parbox{\textwidth}{\raggedright\scriptsize
UTMOS is full band in all rows we ran. Open means the weights can be downloaded, whatever the license. A dash means not published. No clean-rate difference between Mini 1.0.1 and the rows above Chatterbox is distinguishable. Mini 1.0.1's own-harness row used the final weights with the 1.0.1 EQ passed as options. Appendix~\ref{app:repro} says how these rows reproduce.\par\vspace{2pt}
\textbf{Parameters.} \textsuperscript{a}\,The count leaves out the part that makes audio. \textsuperscript{b}\,The count does not say whether it includes the vocoder, so Figure~\ref{fig:scatter} half-fills its marker. \textsuperscript{c}\,All parts from the file headers; the published 500M matches the T3 transformer alone. \textsuperscript{d}\,About 2.29B in the main checkpoint, about 2.38B with the AudioVAE. Alania-2's 488M and Alania-1's 3.3B are our internal counts. EMA Lightning's card splits its 8.6M into a 5.6M acoustic model and a 3.0M HiFi-GAN-style decoder. Sources disagree on XTTS-v2 and Anka TTS (footnote in Section~\ref{sec:related}).\par\vspace{2pt}
\textbf{Quoted rows.} We did not run ElevenLabs and have no per-sentence data, so no interval exists and we make no claim of a difference. For scale, the gap between v4 and our public-harness values, 0.34 WER points and 0.05 CER points, is about the half-width of a paired interval at this test size. The card states v4's UTMOS only in its text.\par\vspace{2pt}
\textbf{Sources.} Trendyol-TTS \citep{trendyol2026tts}, Anka TTS \citep{krmkayabasi2026anka}, Chatterbox \citep{resemble2025chatterbox}, Piper \citep{hansen2023piper}, XTTS-v2 \citep{casanova2024xtts}, Antalia-1 \citep{saygili2026antalia1}, MMS-TTS \citep{pratap2023mms}.}
\end{table}

\begin{figure}[t]
\widefig{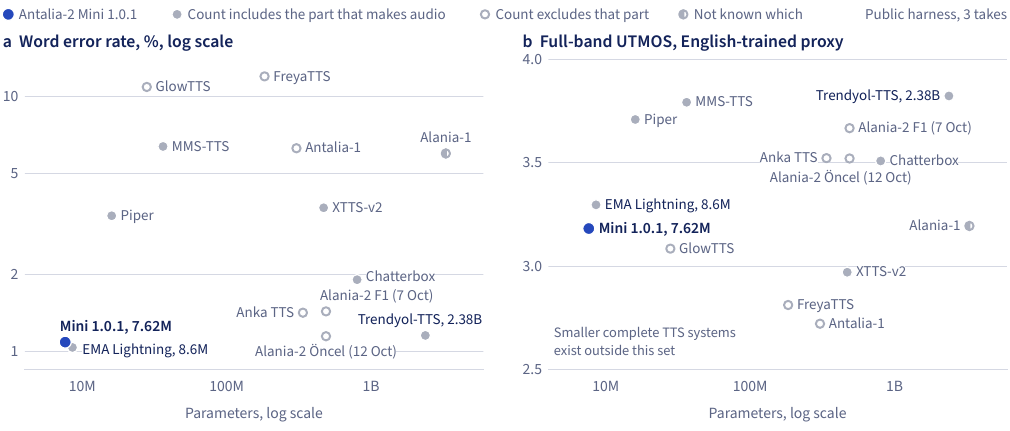}
\caption{\textbf{Mini at 7.62M and Trendyol-TTS at 2.38B are not distinguishable on WER (Figure~\ref{fig:forest}); Mini is lower on full-band UTMOS.} UTMOS is an automatic, English-trained proxy.}\label{fig:scatter}
\end{figure}

Among the 15 systems Mini is the smallest of the 13 with a known size (Figure~\ref{fig:scatter}); some of those counts exclude the vocoder, and Gemini's sizes are not published. Smaller complete TTS systems exist outside this table.

\paragraph{Paired comparison.} Paired bootstrap intervals are the central 95\% of corpus-level differences over at least 10,000 resamples of the 495 sentences, with each sentence's two scores kept together. We call Mini ``level with'' another system when this interval includes 0 on both WER and CER, on the public harness, on our own harness and pooled. The test then cannot tell the two apart; they need not be equal. With 495 sentences and 3 takes the half-width is about 0.27 WER points and 0.07 CER points, so smaller differences cannot be resolved. The term never extends to UTMOS.

Mini 1.0.1 is level with EMA Lightning 1.0.1 (Table~\ref{tab:paired}). Pooled, Mini has 248 word errors to EMA Lightning's 240 of 23,502, and 344 character errors to 342 of 161,844. The pooled result is only a sensitivity check, because own-harness seeds are not paired across harnesses.

\begin{table}[t]
\caption{\textbf{Every paired interval against EMA Lightning 1.0.1 and the teacher includes 0 on WER and CER.} Differences in points, with 95\% intervals. The teacher has one seed, so its intervals resample sentences only.}\label{tab:paired}
\centering\tabfont
\begin{tabular}{@{}lll@{}}
\toprl
\thd{Mini 1.0.1 minus} & \thd{WER} & \thd{CER} \\
\midrule
EMA Lightning 1.0.1, public harness & +0.05 [$-$0.22, +0.32] & $-$0.005 [$-$0.075, +0.065] \\
EMA Lightning 1.0.1, own harness & +0.02 [$-$0.25, +0.27] & +0.007 [$-$0.052, +0.071] \\
EMA Lightning 1.0.1, pooled over 6 takes each & +0.03 [$-$0.21, +0.28] & +0.001 [$-$0.057, +0.062] \\
VoxCPM2 teacher, own harness, 1 seed & +0.10 [$-$0.25, +0.45] & +0.005 [$-$0.103, +0.105] \\
\botrl
\end{tabular}
\end{table}

\begin{figure}[t]
\widefig{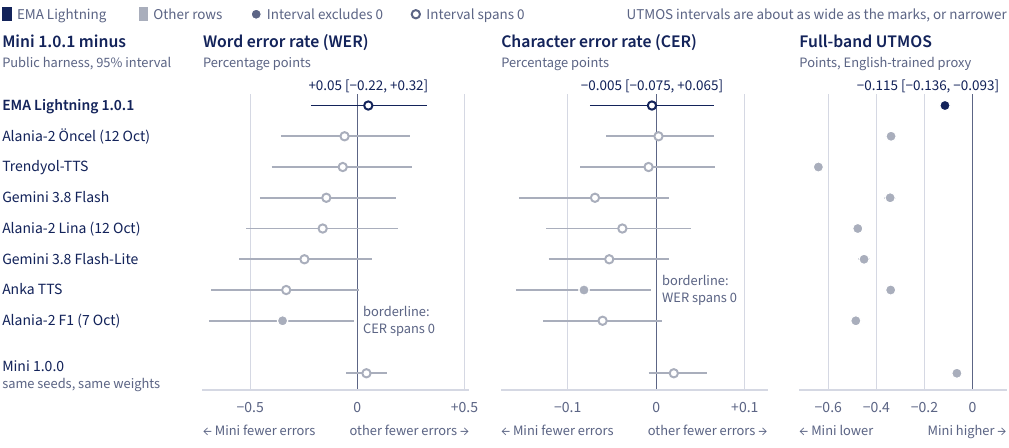}
\caption{\textbf{Mini 1.0.1 is level with EMA Lightning 1.0.1 on WER and CER and lower on full-band UTMOS.} ``Level with,'' never used for UTMOS, means the WER and CER intervals include 0 on both harnesses and pooled.}\label{fig:forest}
\end{figure}

No public-harness WER or CER difference from Alania-2 Öncel or Lina, Trendyol-TTS, or Gemini 3.8 Flash or Flash-Lite is distinguishable (Figure~\ref{fig:forest}). Against Alania-2 F1 (7 Oct) and Anka TTS the intervals exclude 0 on one of WER and CER only, and on neither once takes are resampled as well as sentences, so we call them borderline. Mini has distinguishably fewer word and character errors than the eight systems from Chatterbox down in Table~\ref{tab:results}. In Figure~\ref{fig:forest} the UTMOS intervals all exclude 0.

\paragraph{EMA Lightning as published.} The card \citep{aslan2026emalightning} reports WER 0.92 and CER 0.18, on a release candidate with its own normalizer according to its repository. The pip package 1.0.1 scored 1.00 and 0.198 on our own harness and 1.04 and 0.225 on the public one. PyPI has since moved to 1.1.2, which we did not re-run.

\paragraph{Against the teacher.} On our own harness the student is not distinguishable from VoxCPM2 (1 seed) on WER or CER (Table~\ref{tab:paired}), at about 1/300 of the size. It is lower on UTMOS ($-$0.24) and on 8\,kHz UTMOS ($-$0.20).

\begin{figure}[t]
\widefig{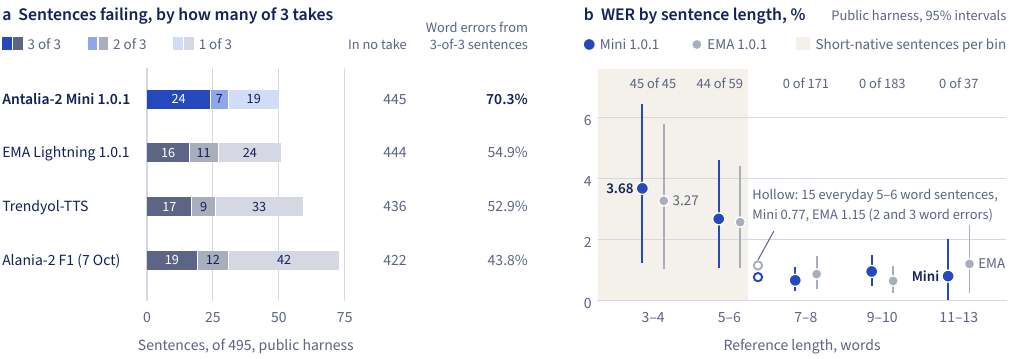}
\caption{\textbf{Mini's errors repeat across takes more than those of the other three systems, and length is confounded with register.} Public harness. (a) Many errors shared with EMA Lightning are recognizer or scorer effects.}\label{fig:persist}
\end{figure}

\paragraph{Where the errors sit.} A sentence that fails in every take points to a cause in the model, the recognizer or the scorer. On the public harness, Mini 1.0.1's errors repeat across takes more than those of EMA Lightning, Trendyol-TTS and Alania-2 F1 (7 Oct) (Figure~\ref{fig:persist}a). A sentence fails in a take if it has at least one word error, and 70.3\% of Mini's word errors sit in sentences that fail in all 3 takes, against 54.9\% for EMA Lightning. Of Mini's 24 such sentences, 17 also fail on all 3 own-harness seeds.

Many shared errors are recognizer or scorer effects. Mini and EMA Lightning both fail 15 sentences in at least 2 of 3 takes, and 12 of these also fail for the teacher. Ten sentences fail that often for Mini and for all eight rows from EMA Lightning to Alania-2 F1 (7 Oct) in Figure~\ref{fig:forest}, costing Mini 0.357 WER points (42 of its 128 word errors) and EMA Lightning 0.349.

Other repeats look like model defects. Sixteen sentences fail that often for Mini but not for EMA Lightning. None of them fails for the teacher, and from transcripts alone, with no listening, 12 look like real sound errors. In one, Whisper transcribes \tr{Çorbanın} as \tr{Şorbanın} in 3 of 3 takes, while 0 of 24 takes of other systems fail.

Short sentences score worse (3 to 4 words: Mini 3.68\%, EMA Lightning 3.27\%), but length is confounded with register, since all of those and 44 of the 59 sentences of 5 to 6 words are short-native. The hollow markers in Figure~\ref{fig:persist}b are the other 15 sentences of 5 to 6 words, all everyday. Mini scores 0.77\% on them (2 word errors), so within that register its short sentences are not worse.

\begin{figure}[t]
\widefig{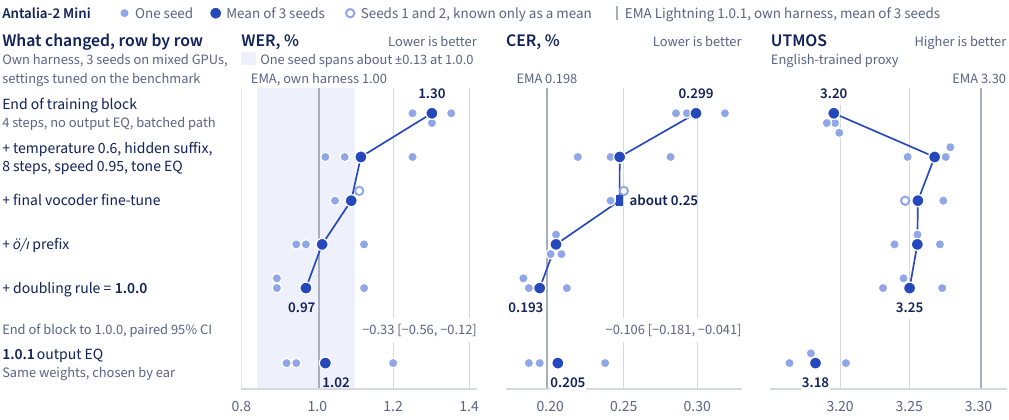}
\caption{\textbf{Settings, two fixes and the final vocoder fine-tune together moved WER from 1.30 to 0.97 and CER from 0.299 to 0.193; no single step is distinguishable on both.} Each row adds to the one above. The row bundling five settings, with the code path, moved WER most, 1.30 to 1.115. Top row: no output EQ, batch 16. Third-row CER is about 0.25 because only a range, 0.2447 to 0.2496, survives for that run.}\label{fig:waterfall}
\end{figure}

\paragraph{How the results moved.} From the end of the 23\,h rental, the training block of Figure~\ref{fig:block}, to 1.0.0, settings, the two fixes of Section~\ref{sec:fixes} and the final vocoder fine-tune together moved Mini from WER 1.30 to 0.97 ($-$0.33 [$-$0.56, $-$0.12]) and from CER 0.299 to 0.193 ($-$0.106 [$-$0.181, $-$0.041]) on our own harness, 3 seeds (Figure~\ref{fig:waterfall}). No single step is distinguishable on both metrics, and the UTMOS panel has no paired interval. We tuned these settings on the benchmark, and the first row used another, batched code path.

At the final settings on our own harness, 8 steps gave WER 0.97 and CER 0.193 over 3 seeds, against 1.17 and 0.230 at 4 steps. With the prefix but no doubling rule, 16 steps gave no gain and temperature 0.5 tied 0.6 (Appendix~\ref{app:settings}).

\begin{figure}[t]
\widefig{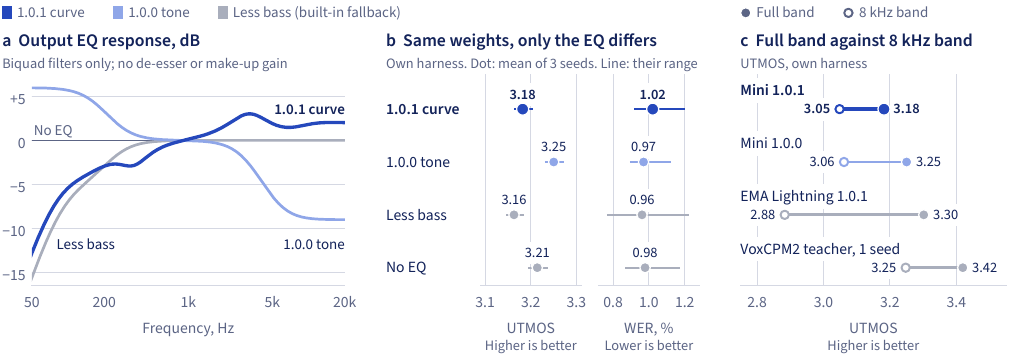}
\caption{\textbf{Automatic scores moved slightly against the 1.0.1 EQ, which we chose by ear.} (b) The 1.0.1 row includes its de-esser and fixed make-up gain. Paired 95\% intervals on our own harness, 1.0.1 minus 1.0.0: UTMOS $-$0.068 [$-$0.074, $-$0.062], WER +0.05 [$-$0.05, +0.16]. (c) Mini is ahead of EMA Lightning in the 8\,kHz band and behind at full band. UTMOS is an automatic, English-trained proxy, and the two models have different voices.}\label{fig:eq}
\end{figure}

\paragraph{UTMOS and the output EQ.} EMA Lightning is ahead on full-band UTMOS on the public harness, 3.30 against 3.18, with Mini minus EMA Lightning at $-$0.115 [$-$0.136, $-$0.093]. The benchmark's card asks for UTMOS in the 8\,kHz band, which only our own harness measures, and there Mini is ahead, 3.05 against 2.88, +0.165 [+0.144, +0.186] (Figure~\ref{fig:eq}c). We have no measurement of why the sign changes. UTMOS is an English-trained proxy and the two models have different voices, so we read both gaps with care (Section~\ref{sec:related}).

We have no record of a listener count or a blind test for the 1.0.1 EQ (Section~\ref{sec:model}). Against 1.0.0, public-harness UTMOS moved by $-$0.065 [$-$0.071, $-$0.060], with WER and CER within noise (last row of Figure~\ref{fig:forest}).

\begin{lesson}\textbf{Lesson.} Freeze the output EQ before using UTMOS to compare checkpoints. On the same weights, four EQ settings span UTMOS 3.16 to 3.25 on our own harness (Figure~\ref{fig:eq}b). With the first vocoder fine-tune, UTMOS rose with more bass and less treble while WER stayed flat (24 runs, seed 0).\end{lesson}

\section{Speed and where it runs}\label{sec:speed}
\standfirst{On one NVIDIA L4 (all 495 sentences, different VMs) Mini 1.0.1 runs at 35\texttimes{} real time, 1.65 times slower than plain EMA Lightning 1.0.1 by RTF and 9.5 times slower than compiled. On two threads of an Apple M5 laptop CPU (first 100 sentences, other jobs running) it runs at 41\texttimes{}. No real phone has been measured.}
\suppressfloats[t]

\begin{figure}[t]
\widefig{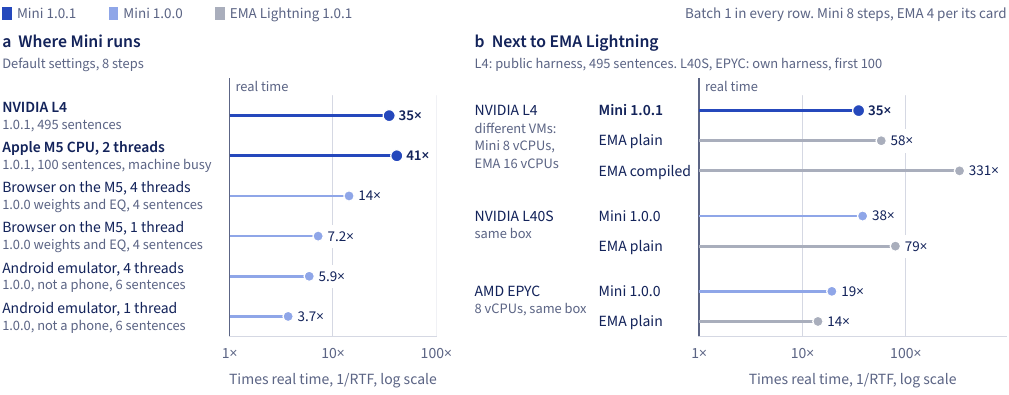}
\caption{\textbf{EMA Lightning was faster on both GPUs compared, and Mini 1.0.0 on the one CPU compared; no real phone was measured.} The L4 row covers all 495 sentences and the M5 row the first 100, so the two do not rank the machines. The 1.0.0 L40S and EPYC runs rebuilt the EQ filter per sentence; 1.0.1 was not timed there. The M5 CPU test had other jobs running.}\label{fig:speed}
\end{figure}

\paragraph{GPU.} On one NVIDIA L4, Mini 1.0.1 has an RTF of 0.029 (35\texttimes{} real time). EMA Lightning 1.0.1 has 0.017 (58\texttimes{}) in plain PyTorch and 0.0030 (331\texttimes{}) compiled, with the same WER and CER (Figure~\ref{fig:speed}b). By wall time the gap to plain EMA Lightning is 1.87 times, because Mini speaks 13\% longer on the same text.

The public harness reports the middle of 3 passes over all 495 sentences at batch 1, with the normalizer and the output chain inside the timed region. Mini samples 8 steps and EMA Lightning 4, per its card \citep{aslan2026emalightning}. The VMs differ, with 8 vCPUs for Mini and 16 for EMA Lightning. The timed render uses the published wheel's cached EQ filter; 1.0.0, without that cache, has an RTF of 0.043. The public-harness scores in Section~\ref{sec:eval} come from an earlier pass with the same seeds and audio length, and we did not compare the audio.

\paragraph{Laptop CPU.} On an Apple M5 CPU with 2 threads, the package default, and 8 steps, version 1.0.1 has an RTF of 0.024 (41\texttimes{}) on the first 100 benchmark sentences (Figure~\ref{fig:speed}a). This is a median of 4 passes with other jobs running, at load 8 to 12 on 10 cores, and these runs of 5 November 2026 are the only ones with recorded machine load. On the model card's sample text, the first chunk from \code{stream()}, the package's streaming call, held 2.8\,s of audio and took 72\,ms. The L4 run covers all 495 sentences, so the two runs do not rank the machines. We have no EMA Lightning number on this machine.

\paragraph{Same machine as EMA Lightning.} On one AWS g6e.2xlarge, Mini 1.0.0 was 2.07 times slower than plain EMA Lightning 1.0.1 on the L40S (RTF 0.026 against 0.0126) and 1.36 times faster on the AMD EPYC 7R13 with 8 vCPUs (0.052 against 0.071; Figure~\ref{fig:speed}b). Our own harness ran the first 100 sentences, one pass, seed 0. Version 1.0.0 rebuilt its EQ filter per sentence, which alone took the L40S from 0.012 to 0.029 over 495 sentences. Version 1.0.1 caches the filter and has not been re-timed on this machine, so its ratios here are unknown. We did not run EMA Lightning's compiled path here.

\begin{table}[t]
\caption{\textbf{The package fetches 30.5\,MB of weights on first use, the browser page 22.8\,MB of graphs and a 14.2\,MB runtime, and the arm64 Android package (APK) is 35.5\,MB.} The weight files and browser graphs are identical in 1.0.0 and 1.0.1. The fp16 file is converted back to fp32 at load, and its rounded durations can change on some sentences. The four ONNX graphs hold the networks, and the page's JavaScript does the iSTFT, duration rounding, noise, chunking, cuts, EQ and limiter.}\label{tab:artifacts}
\centering\tabfont
\begin{tabular}{@{}lrl@{}}
\toprl
\thd{File} & \thd{Bytes} & \thd{Holds} \\
\midrule
\us Weights, fp32, default download & 30,499,764 & Acoustic 14,757,148; vocoder 15,742,616 \\
Weights, fp16, \code{variant="fp16"} & 15,257,774 & Acoustic 7,383,130; vocoder 7,874,644 \\
\midrule
Python wheel, 1.0.1 & 1,397,120 & Inference code and normalizer data, no weights \\
\midrule
Browser graphs, mixed, page default & 22,848,212 & Text, condition, flow in fp32; vocoder in fp16 storage \\
WebAssembly (WASM) binary & 14,239,897 & ONNX Runtime Web 1.30.0 \\
\midrule
Android APK, arm64-v8a & 35,549,175 & Engine v0.1.1, model 1.0.1 \\
Android APK, armeabi-v7a & 34,574,333 & Same engine, 32-bit ARM \\
\botrl
\end{tabular}
\end{table}

\paragraph{Browser.} The browser page, a Hugging Face Space, runs the model client-side with ONNX Runtime Web \citep{onnxruntime}. It has run version 1.0.1 since 6 November 2026, and Table~\ref{tab:artifacts} gives its files. It matches the Python package to 0.036\,dB mean log-spectral distance, the frame-averaged gap between two spectra, on 50 sentences. Lengths are identical, and so are Whisper transcripts on 50 of 50. As a control, a different seed gives 48 of 50.

We timed the page at 1.0.0, with the same weights and graphs and 1.0.0's EQ. In headless Chrome on the same M5, with WebAssembly, it took about 0.38\,s per demo sentence of 4.2 to 7.3\,s on 4 threads (14\texttimes{}) and ran at 7.2\texttimes{} on 1 thread. We have not re-timed it at 1.0.1, whose output chain costs 4.8\,ms per second of audio in Node on the M5, against 0.42\,ms for 1.0.0's. At released settings we have not measured Safari, Firefox, phone browsers, WebGPU or a first download over a real network.

\paragraph{Android.} A signed Android system TTS engine (v0.1.1, sideload, Android 7.0+, arm64 and 32-bit ARM) is publicly downloadable. It runs model 1.0.1, with an app shell adapted from sherpa-onnx \citep{sherpaonnx}. On an Android API 35 arm64 emulator, a debug build matches the package at 1.0.1 to 0.036\,dB on the same 50 sentences, with identical Whisper transcripts on 50 of 50. As a control, a different seed, checked on 10 sentences, differs by 3.74\,dB. The signed arm64 APK itself loaded 1.0.1 and spoke in a smoke test there, and it has no network permission. The 32-bit APK has not been run.

With 4 vCPUs on the M5, that emulator ran a non-debuggable build of the earlier engine v0.1.0, with 1.0.0's output chain, at 3.7\texttimes{} on 1 thread and 5.9\texttimes{} on 4 threads at 8 steps. At 1.0.1 the chain took 10.5\,ms per second of audio on the debug build at 2 threads, 6.9\% of that run's time, and we have not re-timed the emulator's speed. Our test notes put run-to-run noise at up to about 30\%.

\paragraph{Package and API.} The Python package installs from PyPI \citep{antalia2026mini}.

\begin{codeblock}\begin{alltt}
pip install antalia-mini

from antalia_mini import Antalia
tts = Antalia()   # CUDA if available, else CPU; fetches the fp32 weights once
text = "Randevunuz 14 Ekim Salı günü saat 15:30\textquotesingle{}da."
speech = tts.say(text, path="randevu.wav", seed=0)   # 48 kHz by default
for chunk in tts.stream(text, seed=0):               # first sentence first
    ...
\end{alltt}\end{codeblock}

The model is also free on our hosted API, with an API key and limits of 60 requests a minute and 5,000 a day. It is served on CPU at 1.0.1, and its latency is unmeasured. Table~\ref{tab:open} lists what is public, the Availability section gives the links, and Appendix~\ref{app:repro} says what a reader can re-run.

\begin{table}[t]
\caption{\textbf{Weights, inference code and the benchmark harness are public; training code and training data are not.} Status on 6 November 2026. The first four public items are under Apache-2.0. The wheel's \code{model.py} holds the training forward pass but imports a module the wheel does not ship, so the wheel does not train.}\label{tab:open}
\centering\tabfont
\begin{tabular}{@{}ll@{}}
\toprl
\thd{Public} & \thd{Not public} \\
\midrule
Weights, fp32 and fp16 (Hugging Face) & Training code \\
Inference code, Python normalizer (PyPI wheel, sdist) & Data generation scripts \\
JavaScript normalizer, browser page, ONNX graphs (Space) & Training data, the chosen reference clip \\
Benchmark harness, adapters, per-clip scores (Hugging Face) & Own-harness code \\
Signed Android APK, binary only & ONNX export and verify tools, parity tests \\
Hosted API, needs a key & Android source, Kotlin normalizer source \\
 & API server code \\
\botrl
\end{tabular}
\end{table}

\section{Related work}\label{sec:related}

\standfirst{Training one model on another model's audio has precedents back to at least 2021. We found no paper that uses shortcut self-consistency in a TTS acoustic model, and the one Turkish human MOS study of open systems we found orders them differently from our UTMOS, on different renders.}

\paragraph{Synthetic training data.} The earliest precedent we found is TTS-by-TTS \citep{hwang2021ttsbytts}, which added 179\,h of speech from a Tacotron 2-based teacher to 5\,h of recordings. It reports a MOS of 3.74, 40\% above the conventional method. \citet{zhou2025purelysynthetic} train a 20.9M Matcha-TTS student on 100\,h of purely synthetic speech.

\begin{figure}[ht]
\widefig{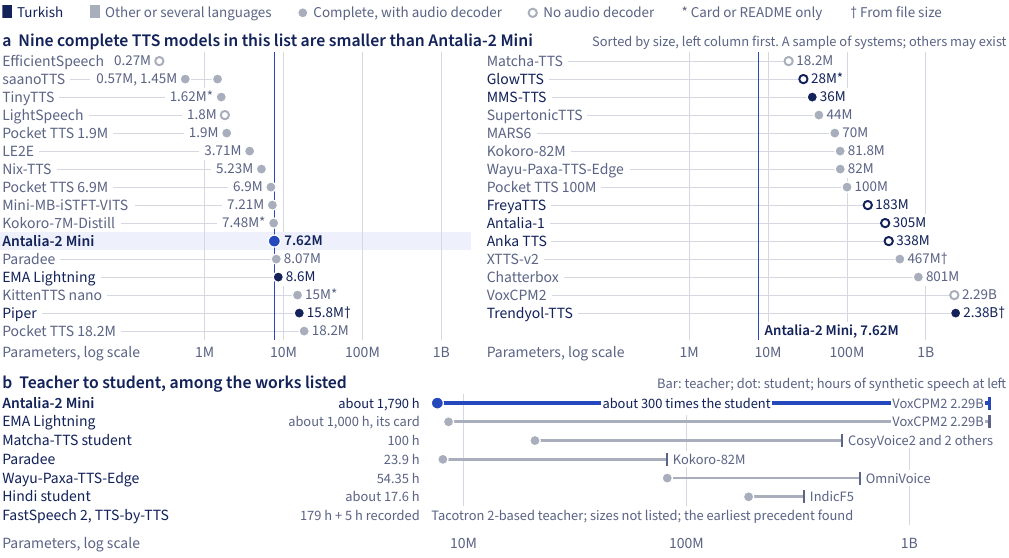}
\caption{\textbf{Nine complete TTS models in this list are smaller than Mini; among the works listed, Mini used the most synthetic speech, about 1,790\,h.} (a) Parameter counts as reported, not like for like; saanoTTS counts twice; Chatterbox at its full 801M. (b) VoxCPM2 is 2.29B, or 2.38B with its AudioVAE. Matcha-TTS's bar marks CosyVoice2, though that work also used XTTS-v2 and ChatTTS. EMA Lightning's row follows its card \protect\citep{aslan2026emalightning}; others, their sources \protect\citep{zhou2025purelysynthetic,mahendrakar2026paradee,pipatanakul2026thai,trikutam2026hindi,hwang2021ttsbytts}.}\label{fig:related}
\end{figure}

Three recent works come closest to ours. By its card, EMA Lightning shares our teacher model (Section~\ref{sec:intro}). Wayu-Paxa-TTS-Edge \citep{pipatanakul2026thai} is a fixed-voice Thai student trained only on synthetic speech from a 15\,s reference. Its authors state that teacher errors become training targets, which matches the start-of-clip \tr{ö} defect our student learned (Section~\ref{sec:fixes}). Paradee \citep{mahendrakar2026paradee} is an 8.07M single-voice English student of Kokoro-82M \citep{hexgrad2025kokoro}. It also uses the teacher's internal features, so it learns no alignment, while we train on audio alone.

\citet{trikutam2026hindi} reports that a small Hindi model trained from scratch on about 17.6\,h fails, and warm-starts the 190M student in the figure from its teacher. \citet{geng2026bridging} find that synthetic data improves phonetic accuracy while suppressing prosodic variability, the range of rhythm and pitch. We did not check our model for this.

\paragraph{Small models.} Figure~\ref{fig:related}a, a sample of systems, lists nine complete models smaller than Mini. They are two saanoTTS models \citep{thapa2026saanotts}, two Pocket TTS models \citep{orsini2026tinytts}, LE2E \citep{vecino2023le2e}, Nix-TTS \citep{chevi2022nix}, Mini-MB-iSTFT-VITS \citep{kawamura2023mbistft} and, by their own README and card, TinyTTS \citep{tinytts2026} and Kokoro-7M-Distill \citep{oddadmix2026kokoro7m}.\footnote{SupertonicTTS is 44M in its report and 66M on its released card \citep{kim2025supertonic,supertone2025supertonic}; we plot 44M. Paradee is 8.07M in its paper and 7.88M in its later README; we plot the paper. XTTS-v2 is 467M by file size, 470M on cards and 750M in \citet{zhou2025purelysynthetic}; we plot file size. Anka TTS is 336M on its card and 338M by our count of its files; we plot the count. Matcha-TTS is 18.2M in panel a \citep{mehta2024matcha} and 20.9M in panel b, from Zhou et al.} The figure leaves out speed, because hardware differs between sources, and quality, because UTMOS across languages and voices is not comparable.

\paragraph{Few-step sampling.} We use shortcut self-consistency \citep{frans2025shortcut}. Its original paper evaluated it on images, and Flow2GAN tested it for mel-conditioned vocoding \citep{yao2026flow2gan}. We found no paper that uses it in a TTS acoustic model, which is absence of evidence only. EMA Lightning's card describes another route, a four-step student distilled with DMD2 \citep{yin2024dmd2} from a same-size flow teacher that needs 16 midpoint steps. \citet{nguyen2025ism} report image artifacts when classifier-free guidance meets shortcut models. Our generator combines the two, and nothing in our evaluation isolates this effect.

\paragraph{Vocoder fine-tuning.} Tacotron 2 \citep{shen2018tacotron2} trains its WaveNet vocoder on ground-truth-aligned predictions and reports a MOS of 4.526 when synthesizing from predicted features, against 4.362 for a vocoder trained on ground truth. HiFi-GAN \citep{kong2020hifigan} fine-tunes on predicted mels and reports MOS 3.91 to 4.18 for its V1, 3.88 to 4.12 for V2 and 3.89 to 4.02 for V3, with mel error rising as perceptual quality rose. Those vocoders train on teacher-forced features. Our mels are sampled, with durations from alignment to the target clip and no teacher forcing, and we score WER, so the effect sizes do not compare. JETS avoids the step by training both stages jointly \citep{lim2022jets}, and \citet{guo2022multiscale} report that a discriminator on the acoustic model is more effective than fine-tuning.

\paragraph{Turkish TTS.} The FreyaTTS report \citep{pamuk2026freyatts} has the only Turkish human MOS of open systems we found, from 24 raters and 2,566 ratings. Among its means are XTTS-v2 3.82, FreyaTTS 3.68, F5-TTS 3.63, MMS-TTS 3.58, Piper 3.47 and GlowTTS 2.53. Mini and EMA Lightning are not in it. Its order of systems disagrees with our UTMOS order on different renders, so we report no correlation. FreyaTTS and, by its card, EMA Lightning also read characters.

\paragraph{Speech quality metrics.} The UTMOS strong learner we use comes from the English main track of the VoiceMOS 2022 challenge \citep{saeki2022utmos}, and we found no validation against Turkish listeners. \citet{asonitis2026limits} find that automatic predictors agree with listeners on audible defects but fall near chance on clean TTS. \citet{takagi2026discrepancies}, on Japanese speech, find them insensitive to prosodic errors, with most scoring lower-F0 speakers higher, and Mini and EMA Lightning have different voices. Paradee's author reports a 0.5 UTMOS gap between two decoders that sounded alike by ear \citep{mahendrakar2026paradee}, and saanoTTS's author reports that aggregate predictors missed a failure on sibilants (s-like sounds) evident in listening \citep{thapa2026saanotts}. \citet{ozkose2026comprehensive} find that rankings of Turkish systems over eighteen objective metrics depend largely on metric family and utterance length.

\section{Limitations and responsible use}\label{sec:limits}

\standfirst{No human listening test of this model exists, and we tuned its settings on the benchmark that scores it. The Whisper checkpoint that scores the benchmark also filtered the training data. The teacher's start-of-clip defect remains in the model, hidden by the prefix, and we have not audited the voice's likeness to any real speaker.}

\paragraph{Evidence.} All our quality numbers are automatic metrics on one benchmark. WER and CER are its band-matched secondary metric, and UTMOS is an English-trained proxy that we compare across different voices. Own-harness seeds ran on different GPUs and EMA Lightning's on an H200. One seed spans about $\pm$0.13 WER points, and the teacher has one seed.

\paragraph{Selection on the benchmark.} Final-settings numbers are optimistic by selection. The doubling rule came from the benchmark's errors, and we screened temperature, steps, speed, suffix and EQ on its 495 sentences. The public harness uses other seeds on the same sentences. The prefix rule's scope came from training-data statistics, but its symptom also appeared on benchmark sentences. A worse benchmark score counted against one teacher reference (Section~\ref{sec:data}).

Whisper ran at beam 1 in QC and at beam 5 in scoring. Clips it could not transcribe left the long set, and our vocoder and settings choices used its WER. WER may therefore flatter the model relative to human intelligibility, by an amount we have not measured.

\paragraph{Errors that remain.} The long set kept 9,271 lines whose first \tr{ö} word is wrong by Whisper, filtered fine-tunes gave no demonstrated gain, and we have not regenerated the data at the source (Section~\ref{sec:fixes}). On the public harness, Mini 1.0.1's errors repeat across takes more than EMA Lightning 1.0.1's, and some of the repeats look like model defects (Section~\ref{sec:eval}). Nobody has listened to these takes.

\paragraph{Scope.} The model has one voice and one style and speaks Turkish only, with no style or emotion control beyond speed. English words are respelled, and whole English sentences are not supported. The model card names short interjections on their own as the weakest area, flat or rushed \citep{antalia2026mini}, and we have no metric for them alone. The vocoder never saw short-clip audio, and Whisper checked only 1.7\% of the short set. Long text is split at sentence ends into independent chunks, so prosody does not carry across them. Readings follow the normalizer, so Roman numerals, ordinals before names, unusual formats and some foreign names can be read wrongly (Section~\ref{sec:model}).

\paragraph{Speed and builds.} We have measured no real phone, only an Android emulator, and Section~\ref{sec:next} lists the other timing runs we have not made. Other jobs were running during the M5 measurements. Version 1.0.1's default EQ, which we chose by ear, lowers UTMOS by 0.065 against 1.0.0 on the public harness and raises WER and CER within noise.

\paragraph{Retraining, voice and use.} Third parties cannot retrain the model (Appendix~\ref{app:repro}). Section~\ref{sec:data} gives the voice's lineage. We ask users to tell listeners when they are hearing a synthetic voice, and not to use the model to impersonate real people or for fraud, harassment or deceptive calls.

\section{What stays open}\label{sec:next}
\standfirst{Four items stay open: \tr{ö} clips regenerated by the teacher, a human listening test, real phones, and new timing runs of version 1.0.1.}

\paragraph{Regenerate at the source.} Our 4 renders per teacher configuration (Section~\ref{sec:fixes}) cannot pick a setting that reads first words right, so regeneration starts with more renders, checked by ear as well as by Whisper. Then the acoustic model retrains on the new clips and the vocoder fine-tunes on its new mels, a step neither filtered run completed. The defect is gone if the student, with the prefix off, gets \tr{ö} first words right as often as first words with other capital letters, on a set larger than the benchmark's 4 \tr{ö}-initial sentences.

\paragraph{Listening test.} A blind test with Turkish listeners on Mini 1.0.0 and 1.0.1, EMA Lightning 1.0.1 and the teacher would settle three questions that automatic metrics leave open. Listener transcripts would show whether WER flatters Mini. Ratings would show whether the UTMOS gaps to EMA Lightning, which run in opposite directions at full band and at 8\,kHz, match what people hear across two voices, and how the 1.0.1 EQ rates against 1.0.0's.

\paragraph{Phones.} Our only Android speeds come from an emulator with 1.0.0's output chain (Section~\ref{sec:speed}). The engine's app has a speed test, so running the signed packages on real phones would give the missing numbers, including for the 32-bit package, which has not been run at all.

\paragraph{Re-timing.} Timing the browser page and the emulator at 1.0.1 would replace the numbers we measured with 1.0.0's output chain. Our same-machine comparison with EMA Lightning also used Mini 1.0.0, so 1.0.1's ratios on that machine's L40S and 8-vCPU AMD EPYC are unknown (Section~\ref{sec:speed}). Timing Mini 1.0.1 and EMA Lightning 1.0.1 on one machine would settle them, and we have no such run planned.

\section*{Availability}

Weights, inference code, the Python normalizer, the ONNX graphs and page code, and the benchmark harness are public under Apache-2.0. Training code, data generation scripts, training data, ONNX export tools and Android source are not (Table~\ref{tab:open}). The links below were live on 6 November 2026.

\par\vspace{0.3em}
{\tabfont\begin{tabularx}{\textwidth}{@{}>{\raggedright\arraybackslash}p{47mm}>{\raggedright\arraybackslash}X@{}}
\toprl
\thd{What} & \thd{Where} \\\midrule
Model card and weights & \href{https://huggingface.co/cloud0day3/antalia-mini}{huggingface.co/cloud0day3/antalia-mini} \\
Python package & \href{https://pypi.org/project/antalia-mini/}{pypi.org/project/antalia-mini} \\
Browser page & \href{https://huggingface.co/spaces/cloud0day3/antalia-mini}{huggingface.co/spaces/cloud0day3/antalia-mini} \\
Signed Android engine, binary only & \href{https://speech.patientdesk.ai/en/android}{speech.patientdesk.ai/en/android} \\
Hosted API, free with an API key & \href{https://speech.patientdesk.ai/en/docs/quickstart}{speech.patientdesk.ai/en/docs/quickstart} \\
Benchmark harness, adapters, \texttt{results.json}, per-clip transcripts, scores and timing & \href{https://huggingface.co/datasets/PatientdeskAI/freya-tr-eval-rerun}{huggingface.co/datasets/PatientdeskAI/freya-tr-eval-rerun} \\
Benchmark results page & \href{https://speech.patientdesk.ai/en/benchmarks}{speech.patientdesk.ai/en/benchmarks} \\
\botrl
\end{tabularx}\par}

\section*{Credits}

The benchmark is Freya-TR-Eval by Freya, under CC-BY-4.0 \citep{pamuk2026freyatts,freya2026treval}. Our scoring protocol and transcription recipe follow EMA Lightning's card \citep{aslan2026emalightning} and the \code{eval.py} of its earlier repository, \code{canberkkkkkk/ema-tts}, under Apache-2.0, and \code{score\_norm}, the public harness's text normalization for scoring, is copied from that file. The baseline settings for the cloning systems, including the voice reference clip for Chatterbox and XTTS-v2, come from the Anka TTS report \citep{krmkayabasi2026anka}.

\section*{Competing interests}

The authors work at PatientDesk AI, which sells Alania-2 and Alania-1 and ran the benchmark on both harnesses. Both products appear in Table~\ref{tab:results}, along with Antalia-1, our open-weights release. We tuned Mini's settings on the benchmark's own sentences (Section~\ref{sec:limits}).

\bibliographystyle{plainnat}
\bibliography{refs}
\clearpage
\appendix
\section{A selection of the settings we tried}\label{app:settings}

\standfirst{Seventeen of the 31 comparisons here rest on one seed. Of the 14 with 3 seeds, only 8 steps against 4 moved WER by more than one seed's span.}

\begin{figure}[ht]
\widefig{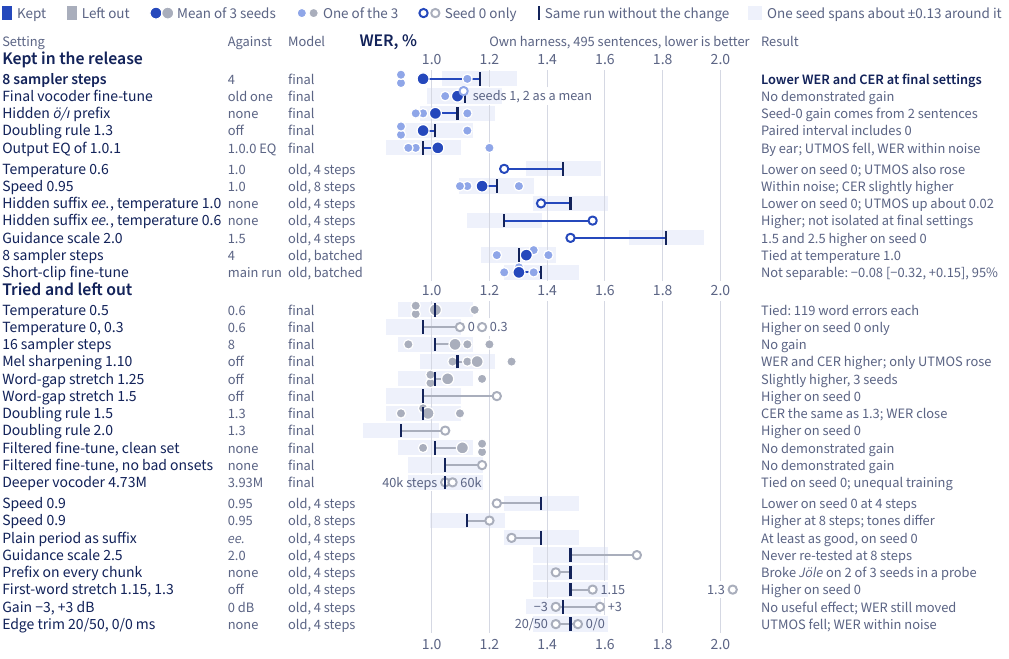}
\caption{\textbf{Of 14 rows with 3 seeds, only 8 steps against 4 at the final settings moved WER past the $\pm$0.13 points one seed spans.} Final: released weights, 1.0.0's EQ unless named. Old: the same acoustic model with the first vocoder fine-tune. Batched: batch 16, temperature 1.0, no suffix or EQ.}\label{fig:settings}
\end{figure}

Figure~\ref{fig:settings} shows about one comparison per setting, picked from 123 evaluation jobs. Each row changes the named setting, on our own harness. In the first row the small dots are the 3 seeds at 8 sampler steps, the large dot is their mean and the dark tick is the same run without the change, at 4 steps. These scores are optimistic by selection, because we tuned on the same 495 benchmark sentences (Section~\ref{sec:limits}). Kept means the release ships the setting, and several kept rows show no gain on WER. Mel sharpening scales each mel bin's variation around its mean, and edge trim cuts silence at chunk edges down to the named milliseconds.

The figure leaves out the old model's temperatures 0.8, 0.7, 0.5 and 0.4, the other suffix variants at temperature 1.0, the other every-chunk prefixes, the 24-run tone sweep and the other EQ curves, and intermediate checkpoints.

The tinted band is one seed's span, applied here to 3-seed means as well as single seeds. It gives a scale only, and a row inside it may still have an effect. Some values come from printed scorer lines instead of per-clip files. On the final vocoder fine-tune row, we kept seeds 1 and 2 only as one printed mean, drawn as one hollow mark.

\section{Reproducing the numbers}\label{app:repro}
\standfirst{The public-harness rows of the open-weights systems can be re-run from public files. Retraining the model or re-running our own harness needs files we have not published.}

\begin{table}[t]
\caption{\textbf{The shipped weights come from 471,880 acoustic updates and 535,350 vocoder updates.} All stages ran on H200 GPUs except the final vocoder fine-tune, on one L40S and then 4 A100 80GB.}\label{tab:checkpoints}
\centering\tabfont
\makebox[\textwidth][c]{\begin{tabular}{@{}lrll@{}}
\toprl
\thd{Stage} & \thd{Updates} & \thd{Starts from} & \thd{Trains on} \\
\midrule
\multicolumn{4}{@{}l}{\thd{Acoustic model, 3,686,977 parameters}} \\
Main run & 420,000 & No parent & Long set \\
Short-clip fine-tune & 51,880 & Main run, step 420,000 & Both sets, short clips \texttimes{}6 \\
\us Shipped acoustic model & 471,880 & & \\
\midrule
\multicolumn{4}{@{}l}{\thd{Vocoder, 3,933,954 parameters}} \\
Real-mel vocoder & 157,000 & No parent & Long set, real mels \\
First vocoder fine-tune & 301,490 & Real-mel vocoder, step 157,000 & Long set, mels from the main run \\
Final vocoder fine-tune & 76,860 & First vocoder fine-tune, step 301,490 & Long set, mels from the short-clip fine-tune \\
\us Shipped vocoder & 535,350 & & \\
\botrl
\end{tabular}}
\end{table}

\paragraph{What a reader can re-run.} The benchmark sentences are Freya-TR-Eval \citep{pamuk2026freyatts,freya2026treval}, and our dataset holds the harness, one adapter per open system, \code{results.json}, and per-clip transcripts, scores and timing \citep{patientdesk2026rerun}. Every one of the 20 rows of \code{results.json}, closed systems included, can be recomputed with zero difference from the per-clip files, without rendering anything. Table~\ref{tab:results} leaves out EMA Lightning's compiled build, Alania-2 F1 (7 Oct) as a bare model, and a 100-sentence Alania-1 speed row. Mini 1.0.0 ran on the same harness and seeds but is not in that file, which now holds 1.0.1. To render again, install package version 1.0.1, run \code{run\_antalia\_mini.py} on a CUDA GPU, then \code{score.py} and \code{aggregate.py}, and compare with \code{results.json}. For EMA Lightning, install 1.0.1.

Table~\ref{tab:checkpoints} traces the shipped weights. On the model repository, commit \code{8a79528} is 1.0.1 and commit \code{e0d3182b} is 1.0.0, and \code{Antalia(revision=...)} selects one. Re-timing 1.0.0 also needs its own wheel from PyPI, because the 1.0.1 wheel caches the EQ filter. Settings are options, so \code{Antalia(prefix="", gem\_stretch=1.0)} turns off the hidden prefix and the doubling rule.

Seeds fix the audio only on one software stack (Section~\ref{sec:model}). Re-scoring the 7 Oct Alania-2 F1 audio on an A100, the GPU that scored the 12 Oct Alania-2 rows, changed 3 of 1,485 transcripts and WER by 0.0085 points. RTF depends on the machine, and ours were L4s on GCP g2-standard-8 for Mini and g2-standard-16 for EMA Lightning.

\paragraph{What a reader cannot re-run.} Training code, data generation scripts, training data and the chosen reference clip are not public (Table~\ref{tab:open}), and the shipped runs used flags that exist only in our private repository, so third parties cannot retrain the model. The data recipe exists only as prose, on the model card and in Section~\ref{sec:data}. The analyses in Sections~\ref{sec:data} to~\ref{sec:fixes} rest on that private material. The own-harness rows, including every 8\,kHz UTMOS value, use code that is not public, and so do the parity tests of Section~\ref{sec:speed} and the normalizer checks below. The dataset has the per-clip transcripts behind the public-harness intervals of Section~\ref{sec:eval} but no bootstrap script, so a reader has to write one.

A normalizer replay of 31,584 recorded text cases on the 1.0.1 code gave 0 mismatches. The browser page's JavaScript port matches the Python code on 1,171 of 1,171 sentences, and the Android engine's Kotlin port on 31,584 of 31,584 cases on the JVM. Replays catch code changes and cannot catch a wrong reading.

\end{document}